\documentclass{article}

\usepackage[preprint]{neurips_2023}

\usepackage[utf8]{inputenc} 
\usepackage[T1]{fontenc}    
\usepackage{hyperref}       
\usepackage{url}            
\usepackage{booktabs}       
\usepackage{amsfonts}       
\usepackage{nicefrac}       
\usepackage{microtype}      
\usepackage{xcolor}         

\usepackage{multirow}
\usepackage{color}
\usepackage{colortbl}
\usepackage{stfloats}
\usepackage{comment}
\usepackage{proof}
\usepackage{fancybox}
\usepackage[flushleft]{threeparttable}
\usepackage{wrapfig}
\usepackage{pgfplotstable} 
\pgfplotsset{compat=1.12}
\usepackage{tabularx}
\usepackage{arydshln}
\usepackage{hyperref}
\usepackage[most]{tcolorbox}
\usepackage{subfig}
\usepackage[noend]{algpseudocode} 

\usepackage{bm}
\usepackage{mathtools}
\usepackage{stmaryrd}
\usepackage{graphicx}
\usepackage{color}
\usepackage{colortbl}
\usepackage{stfloats}
\usepackage{comment}
\usepackage{proof}
\usepackage{fancybox}
\usepackage[flushleft]{threeparttable}

\usepackage{paralist, tabularx}
\usepackage{wrapfig}

\usepackage{pgfplotstable} 
\pgfplotsset{compat=1.12}

\usepackage{tabularx}

\usepackage{arydshln}
\usepackage{hyperref}

\usepackage{pythonhighlight}
\lstnewenvironment{mypython}[1][]{\lstset{style=mypython,#1}}{}

\title{Self-Refined Large Language Model as Automated Reward Function Designer for Deep Reinforcement Learning in Robotics}

\author{%
  Jiayang~Song$^\dagger$ \\\
  University of Alberta\\
  \texttt{jiayan13@ualberta.ca} \\
  \And
   Zhehua~Zhou$^\dagger$ \\\
  University of Alberta\\
  \texttt{zhehua1@ualberta.ca} \\
  \And
  Jiawei~Liu \\\
  Nanjing University\\
  \texttt{jw.liu@smail.nju.edu.cn} \\
  \And
  Chunrong~Fang \\\
  Nanjing University\\
  \texttt{fangchunrong@nju.edu.cn} \\
  \And
  Zhan~Shu \\\
  University of Alberta\\
  \texttt{zshu1@ualberta.ca} \\
  \And
  Lei~Ma \\\
  The University of Tokyo \\
  University of Alberta\\
  \texttt{ma.lei@acm.org} \\
}

\begin{document}

\maketitle

\begin{abstract}
Although Deep Reinforcement Learning (DRL) has achieved notable success in numerous robotic applications, designing a high-performing reward function remains a challenging task that often requires substantial manual input.
Recently, Large Language Models (LLMs) have been extensively adopted to address tasks demanding in-depth common-sense knowledge, such as reasoning and planning.
Recognizing that reward function design is also inherently linked to such knowledge, LLM offers a promising potential in this context.
Motivated by this, we propose in this work a novel LLM framework with a self-refinement mechanism for automated reward function design.
The framework commences with the LLM formulating an initial reward function based on natural language inputs. 
Then, the performance of the reward function is assessed, and the results are presented back to the LLM for guiding its self-refinement process.
We examine the performance of our proposed framework through a variety of continuous robotic control tasks across three diverse robotic systems. 
The results indicate that our LLM-designed reward functions are able to rival or even surpass manually designed reward functions, highlighting the efficacy and applicability of our approach.
All codes and results relevant to this paper are available at \url{https://github.com/zhehuazhou/LLM_Reward_Design}.
\end{abstract}

\section{Introduction}
\label{sec.intro}

Over the past years, substantial progress has been achieved in leveraging Deep Reinforcement Learning (DRL) to tackle a broad spectrum of complex challenges across diverse robotic domains, such as manipulation~\cite{nguyen2019review}, navigation~\cite{zhu2021deep}, locomotion~\cite{yue2020learning}, and aerial robotics~\cite{azar2021drone}.
However, despite these advancements, training high-performing DRL agents remains a challenging task~\cite{andrychowicz2020matters}. 
A principal contributing factor to this complexity is the inherent difficulty in designing an effective reward function, 
which is vital and fundamental to DRL approaches~\cite{sutton1998introduction}.

Conventional methods of reward function design predominantly rely on meticulous manual crafting~\cite{eschmann2021reward}. 
Recent research has introduced Automated Reinforcement Learning (AutoRL) approaches~\cite{parker2022automated}, aiming to automate the hyperparameters and reward function tuning in DRL. 
These approaches commence with a predefined, parameterized reward function and subsequently fine-tune its parameters to identify an optimal reward function~\cite{chiang2019learning,faust2019evolving}. 
However, instead of developing the reward function from scratch, AutoRL remains dependent on an initial parameterized function provided by human experts. 
The construction of such a function often demands domain-specific expertise and a significant investment of time and effort.

In recent research, Large Language Models (LLMs) have been increasingly utilized for tasks that demand common-sense reasoning and extensive world knowledge~\cite{bommasani2021opportunities}, spanning domains like natural language processing~\cite{brown2020language}, task planning~\cite{ahn2022can}, and reasoning~\cite{zelikman2022star}. 
The compelling outcomes from these studies reveal the ability of LLMs to emulate human cognitive processes and integrate a substantial degree of common-sense knowledge~\cite{petroni2019language, davison2019commonsense}.
Given that designing reward functions often also depends on such knowledge, researchers are currently exploring the potential of LLMs as reward function designers for DRL.
Leveraging natural language instructions as input, LLMs are able to formulate effective reward functions for simple tasks in game environments with discrete action spaces~\cite{kwon2023reward}. 
Moreover, their rich internalized knowledge about the world also aids in comprehending both user preferences and task requirements.
However, as LLMs are essentially engineered to generate word sequences that align with human-like context, their efficacy and reliability in reward function design remain uncertain, especially for robotic control tasks that involve continuous action spaces.

\begin{figure*}
    \centering
    \includegraphics[width=\linewidth]{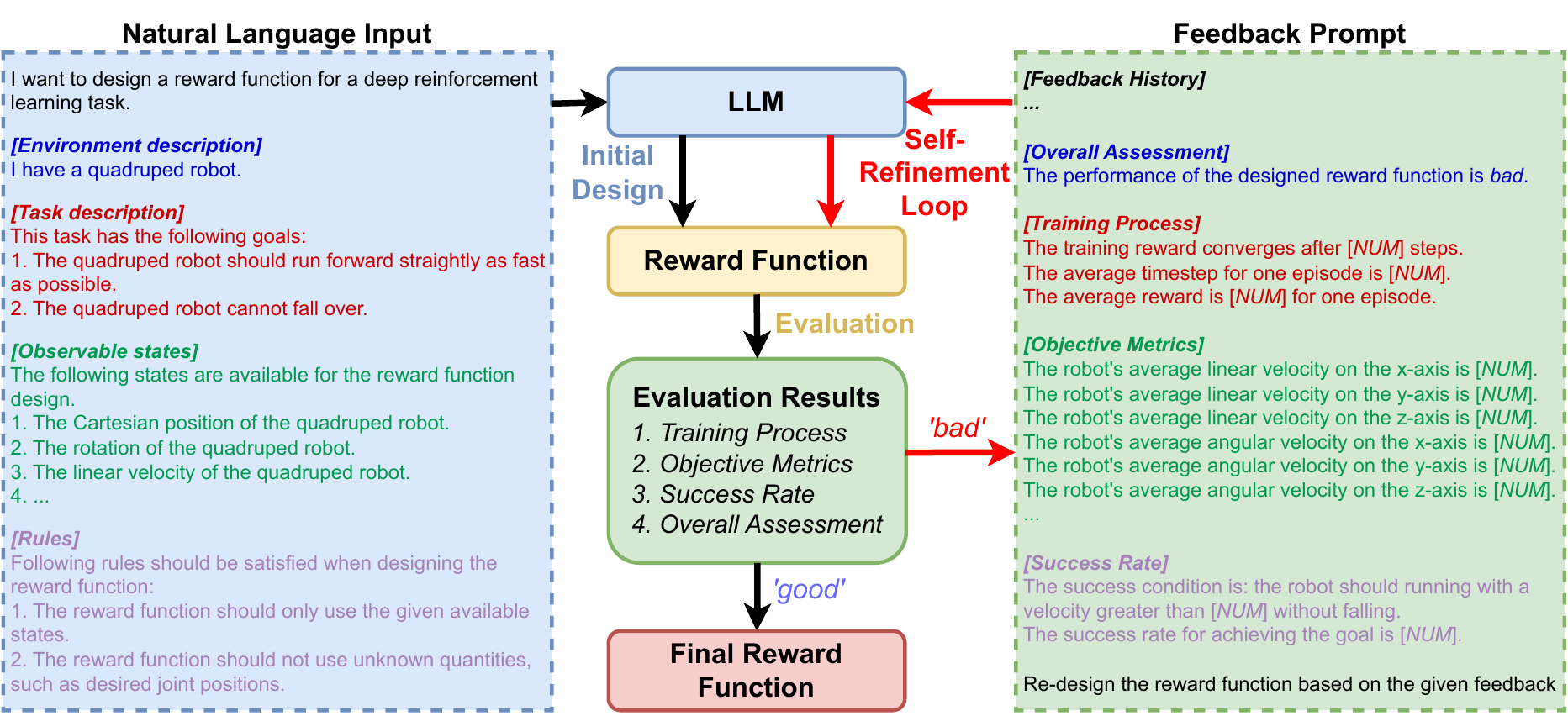}
    \caption{Our proposed self-refine LLM framework for reward function design.
    It consists of three steps: \emph{initial design}, \emph{evaluation}, and \emph{self-refinement loop}.
    A quadruped robot forward running task is used as an example here. 
    A complete list of the prompts used in this work can be found in the appendix.}
    \label{fig.overview}
    \vspace{-10pt}
\end{figure*}

In this work, we investigate the possibility of employing LLM as an automated reward function designer for DRL-driven continuous robotic control tasks.
Motivated by recent studies that demonstrate the capability of LLM for self-refinement~\cite{madaan2023self, huang2022large}, we propose a novel self-refined LLM framework for reward function design.
The framework consists of three steps (see Fig.~\ref{fig.overview}):
1) \emph{Initial design}, where the LLM accepts a natural language instruction and devises an initial reward function;
2) \emph{Evaluation}, where the system behavior resulting from the training process using the designed reward function is assessed;
3) \emph{Self-refinement loop}, where the evaluation feedback is provided to the LLM, guiding it to iteratively refine the reward function. 
To optimize results, the evaluation and self-refinement steps are repeated until either a predefined maximum number of iterations is reached, or the evaluation suggests satisfactory performance.
We examine the performance of the proposed self-refined LLM framework across nine different tasks distributed among three diverse robotic systems. 
The results show that our approach is capable of generating reward functions that not only induce desired robotic behaviors but also rival or even exceed those meticulously hand-crafted reward functions.

The contributions of this paper are threefold:
\begin{itemize}
    \item We explore the ability of LLM to design reward functions for DRL controllers. 
    Diverging from many studies that leverage few-shot in-context learning when prompting the LLM, we employ the LLM as a zero-shot reward function designer.
    \item We incorporate a self-refinement mechanism into the reward function design process to enhance its outcomes.
    \item We highlight the effectiveness and applicability of our proposed approach through a variety of continuous robotic control tasks across diverse robotic systems.
\end{itemize}

\section{Related Work}
\label{sec.related_work}

\noindent
\textbf{Reward Function Design with AutoRL}
AutoRL extends Automated Machine Learning (AutoML)~\cite{he2021automl} principles to address reinforcement learning challenges. 
Its primary focus is to automate the fine-tuning of both the architectures of neural networks and the hyperparameters of learning algorithms~\cite{parker2022automated}. 
For reward function design, AutoRL utilizes evolutionary algorithms~\cite{real2019regularized} to adjust the parameters of a predefined parameterized reward function, aiming to identify an optimal reward function. 
In~\cite{chiang2019learning}, AutoRL is applied to modify the reward function for a navigation problem, while~\cite{faust2019evolving} further expands the use of AutoRL in reward function optimization to multiple reinforcement learning benchmarks simulated in Mujoco~\cite{todorov2012mujoco}. 
In essence, AutoRL can be considered a reward shaping technique~\cite{ng1999policy}.  
However, due to its dependency on an initially hand-crafted parameterized reward function, AutoRL lacks the ability to formulate a reward function entirely from scratch.

\noindent
\textbf{Reward Function Design with LLM} 
Benefiting from its pre-trained common-sense knowledge, LLM offers the potential to alleviate the human effort required in formulating reward functions.
Recent studies have revealed the capability of LLM in directing reward shaping approaches~\cite{mirchandani2021ella,carta2022eager}. 
In~\cite{yu2023language}, instead of creating a reward function for DRL, LLM is employed to determine objective functions for predefined model predictive controllers. 
State-of-the-art research demonstrates that for simpler tasks, such as normal-form games~\cite{costa2001cognition} with discrete action spaces, LLM can serve directly as a proxy reward function~\cite{kwon2023reward}. 
Through processing natural language instructions, LLM seamlessly integrates task requirements and user preferences into reward functions~\cite{hu2023language}. 
However, whether LLM is able to independently design a reward function from scratch for continuous robotic control tasks remains an open research question.

\section{Preliminary}
\label{sec.preliminary}

\noindent
\textbf{DRL Setup}
We model the DRL problem as a Partially Observable Markov Decision Process (POMDP)~\cite{monahan1982state} $\mathcal{M}(S, O, A, T, R, \gamma)$ with continuous state and action spaces.
Given a state $s\in S$, a DRL agent determines an action $a \in A$. 
The system then transitions to a new state $s'$ according to the transition distribution function $T(s'|s,a)$, which results in an observation $o \in O$.
For a given reward function $R: S \times A \rightarrow \mathbb{R}$, the training process of DRL aims to find a policy $\hat{\pi}_R$ that maximizes the expected cumulative discounted reward
\begin{equation}
\hat{\pi}_R = \arg\max_{\pi} \mathbb{E} \left[ \sum_{t=0}^{\infty} \gamma^t R(s_t, a_t) \right],
\label{POMDP}
\end{equation}
where $\gamma$ is the discount factor.

\noindent
\textbf{Reward Function Design}
While the reward function $R$ provides immediate feedback to the DRL agent, it often lacks human interpretability due to the inherent difficulty in directly associating numerical values with system behaviors.
To analyze the performance of a policy, humans typically evaluate system trajectories $\mathcal{T}$ generated by the trained DRL agent through a performance metric $G(\mathcal{T})$.
For example, in the humanoid walking task, the performance metric $G(\mathcal{T})$ could be the maximum distance the humanoid can travel without falling. 
Therefore, the goal in designing a reward function is to determine an optimal reward function $\hat{R}$ that, after the training process, results in a policy $\hat{\pi}_{\hat{R}}$ that maximizes the performance metric $G(\mathcal{T})$.

\section{Self-Refined LLM for Reward Function Design}
\label{sec.method}

In this section, we introduce a self-refined LLM framework for automated reward function design.
It contains three steps: \emph{initial design}, \emph{evaluation}, and \emph{self-refinement loop}.

\subsection{Initial Design}
\label{sec.initial_design}

We first employ the LLM to formulate an initial reward function based on natural language input.
To enhance the LLM's comprehension of the robotic control task, we segment the natural language prompt into four parts (see Fig.~\ref{fig.overview}):
\begin{itemize}
    \item \textit{Environment description}: we first describe the robotic system we are working with, e.g., a quadruped robot or a 7-DOF manipulator, and provide details regarding the environmental setup;
    \item \textit{Task description}: we then outline the control objectives of the task, along with any existing specific task requirements;
    \item \textit{Observable states}: we also provide a list of the observable states that are available for the reward function design;
    \item \textit{Rules}: finally, we explain the rules that the LLM should follow when designing the reward function. 
    Specifically, we emphasize two rules: first, the reward function should be based solely on the observable states; second, the reward function should exclude elements that are challenging to quantify, such as specific target postures of the quadruped robot.
\end{itemize}

Similar to many hand-crafted reward functions, the initial reward function formulated by the LLM is often given as a weighted combination of multiple individual reward components, i.e., we have $R = \sum_{i=0}^{n} w_i r_i$.
However, the initial weights $w_i$ are usually unreliable and necessitate adjustments.
We address this challenge by using our proposed self-refinement process.

It is worth mentioning that while many studies leverage few-shot in-context learning~\cite{brown2020language} to guide the LLM in generating responses in a desired manner, our approach utilizes the LLM as a zero-shot reward function designer, excluding examples in our prompts. 
The major reason is that, due to the inherent task-specificity of reward function design, finding universally applicable examples for a diverse array of robotic control tasks proves challenging. 
To ensure the performance of the designed reward function, we employ the subsequent evaluation and self-refinement processes.
A complete list of the prompts used in our experiments is available  in Appendix~\ref{sec.appendix_input_prompt}.

\begin{figure*}[!t]
\centering
\subfloat[Ball Catching]{\includegraphics[height= 1.6in, width= 1.75in]{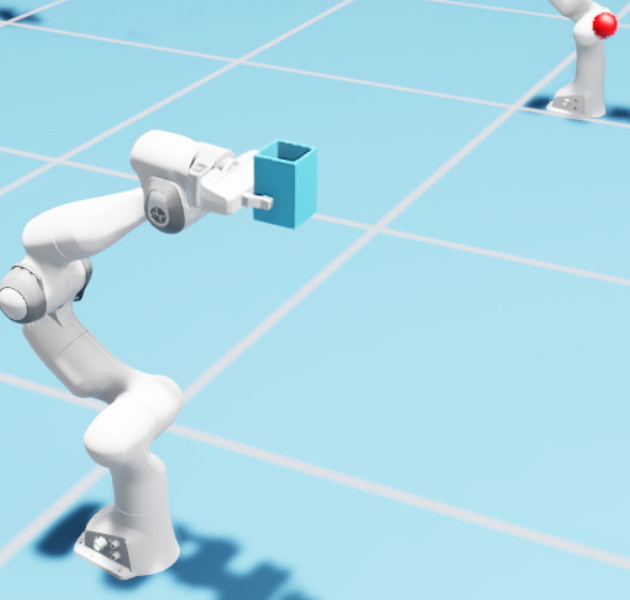}
\label{fig.BC}}
\hfil
\subfloat[Ball Balancing]{\includegraphics[height= 1.6in, width= 1.75in]{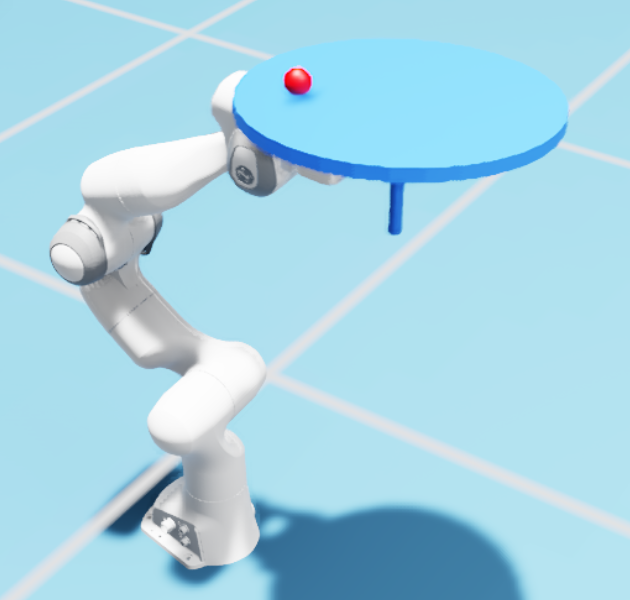}
\label{fig.BB}}
\hfil
\subfloat[Ball Pushing]{\includegraphics[height= 1.6in, width= 1.75in]{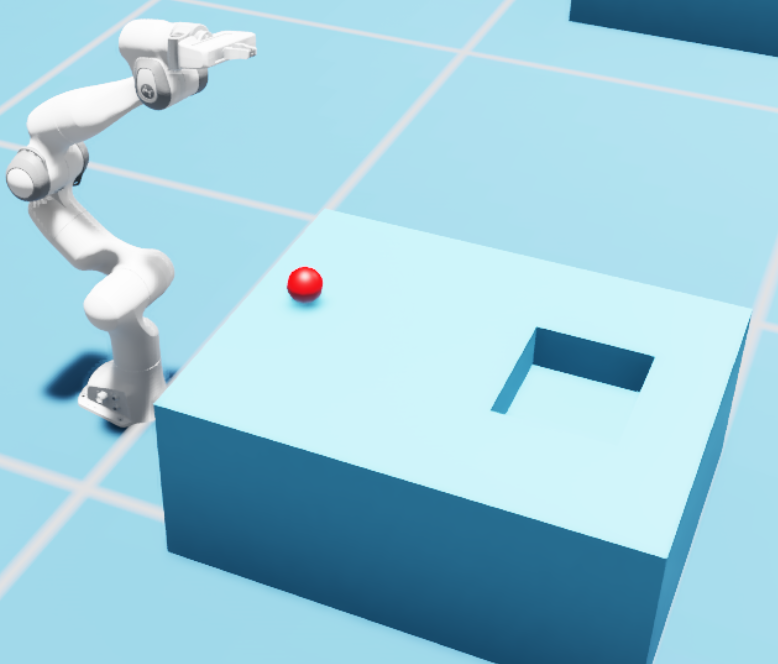}
\label{fig.BP}}
\hfil
\subfloat[Velocity Tracking]{\includegraphics[height= 1.6in, width= 1.75in]{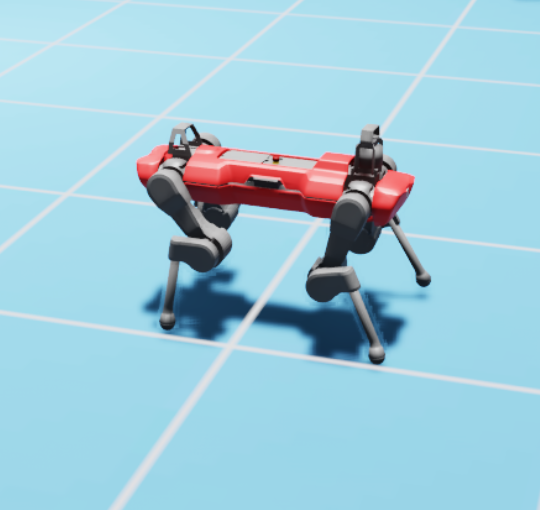}
\label{fig.walking}}
\hfil
\subfloat[Hovering]{\includegraphics[height= 1.6in, width= 1.75in]{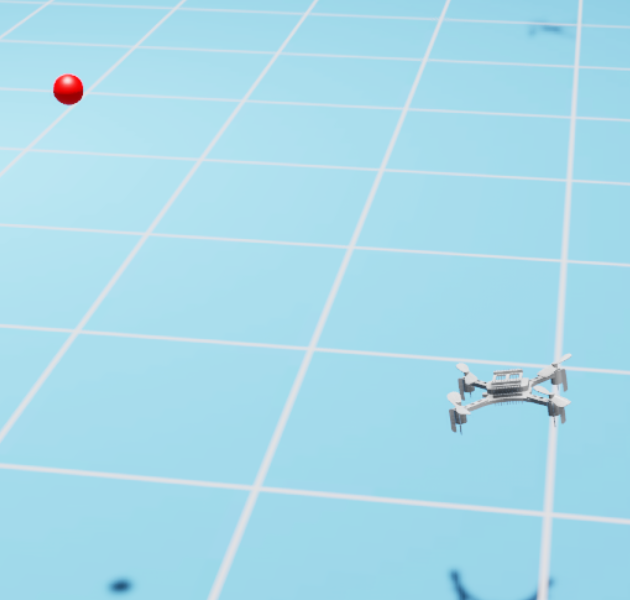}
\label{fig.crazyflie}}
\hfil
\caption{Continuous robotic control tasks with three diverse robotic systems: robotic manipulator (Franka Emika Panda~\cite{franka}), quadruped robot (Anymal~\cite{anymal}) and quadcopter (Crazyflie~\cite{crazyflie}). Simulations are conducted in NVIDIA Isaac Sim~\cite{isaacsim}.}
\label{fig.control_tasks}
\vspace{-10pt}
\end{figure*}

\subsection{Evaluation}
\label{sec.evaluation}

After the LLM determines the reward function $R$, we assess its efficacy via an evaluation process (see Fig.~\ref{fig.overview}). 
Aiming to minimize human intervention, the evaluation is structured as an automated procedure.

We begin by initiating a training process to obtain a trained optimal DRL policy $\hat{\pi}_{R}$. 
Subsequently, we sample $n_t$ trajectories $\mathcal{T}_i, i=1,\ldots,n_t$ of this trained policy $\hat{\pi}_{R}$, each originating from a distinct, randomly selected initial state. 
Performance of the reward function $R$ is then evaluated from the following three aspects:
\begin{itemize}

    \item \textit{Training process:} we first summarize the training process for the policy $\hat{\pi}_{R}$ to evaluate the immediate effectiveness of the designed reward function $R$. 
    This summary includes information on whether the reward has converged, the average reward per training episode, and the average number of timesteps in each episode.
    
    \item \textit{Objective metrics}: we then represent the overarching performance metric $G(\mathcal{T})$ with multiple individual task-specific objective metrics $g_k(\mathcal{T}), k = 1,\ldots,n_g$.
    Each objective metric $g_k(\mathcal{T})$ addresses an aspect of the task requirements.
    For instance, in the quadruped robot's straight-forward walking task, two objective metrics could be employed: one assessing the forward distance the robot travels without toppling and another quantifying any unintended lateral movements. 
    We then compute the average values of these objective metrics $g_k(\mathcal{T})$ over all sampled trajectories.

    \item \textit{Success rate in task accomplishments}: in addition to the task-specific objective metrics, we also introduce the success rate $\mathrm{SR}$ of the trained policy $\hat{\pi}_{R}$ in accomplishing the designated control task as a general and task-agnostic criterion.
    For each control task, we define a success condition using Signal Temporal Logic (STL)~\cite{donze2013signal} to capture the core objective of the task.
    For example, the success condition for a quadruped robot walking task could be that the forward distance travelled without falling should exceed a predetermined threshold.
    A trajectory meeting the success condition is considered a success. 
    The success rate $\mathrm{SR}$ is determined across all sampled trajectories. 

\end{itemize}

As a conclusion of the evaluation, we finally categorize the overall performance of the designed reward function $R$ as either `\emph{good}' or `\emph{bad}'.
Given that the training process and objective metrics are intrinsically task-dependent, it is challenging to establish a universally applicable standard to assess different tasks based on these two criteria.
Therefore, we rely solely on the success rate $\mathrm{SR}$ for the overall assessment and reserve other details as guidance for the subsequent self-refinement process.
If the success rate $\mathrm{SR}$ exceeds a predefined threshold, the performance of the reward function is considered `\emph{good}'. 
Otherwise, we label it as `\emph{bad}' and initiate a self-refinement process to improve.

\subsection{Self-Refinement Loop}
\label{sec.self_refine}

To enhance the designed reward function, we employ a self-refinement process.
It starts with the construction of a feedback prompt for the LLM based on evaluation results (see Fig.~\ref{fig.overview}).
To offer the LLM a clear and immediate comprehension of the performance of the reward function, we position the overall assessment at the beginning of the prompt, followed by detailed information of the training process, objective metrics, and success rate.
Guided by this feedback and previous feedback history, the LLM attempts to develop an updated reward function.
Details about all feedback prompts used in our experiments are presented in Appendix~\ref{sec.appendix_feedback_prompt}.

For finding an optimal reward function, we repeat the evaluation and self-refinement processes in a loop until either a predefined maximum number of iterations is reached, or the evaluation suggests `\emph{good}' performance.
The reward function, resulting from the self-refinement loop, is accepted as the final designed reward function.

\section{Experimental Results}
\label{sec.results}

\subsection{Experimental Setup}
\label{sec.experimental_setup}

We evaluate the performance of our proposed framework in designing reward functions through nine distinct continuous robotic control tasks across three diverse robotic systems (see Fig.~\ref{fig.control_tasks}).
Specifically, we employ the following tasks and systems that are also frequently referenced as benchmark challenges in DRL studies~\cite{james2020rlbench,zhu2020robosuite,zhou2023towards}:
\begin{itemize}
    \item \textit{Robotic manipulator (Franka Emika Panda~\cite{franka})}: 
    \begin{enumerate}
        \item \textit{Ball catching}: the manipulator needs to catch a ball that is thrown to it using a tool (Fig.~\ref{fig.BC});
        \item \textit{Ball balancing}: the manipulator should keep a ball, which falls from above, centered on a tray held by its end-effector (Fig.~\ref{fig.BB});
        \item \textit{Ball pushing}: the manipulator is required to push a ball towards a target hole on a table (Fig.~\ref{fig.BP});
    \end{enumerate}
    \item \textit{Quadruped robot (Anymal~\cite{anymal})}:
    \begin{enumerate}
        \setcounter{enumi}{3}
        \item \textit{Velocity tracking}: the robot needs to walk at a specified velocity without toppling over (Fig.~\ref{fig.walking});
        \item \textit{Running}: the robot should run straight forward as fast as possible without falling;
        \item \textit{Walking to target}: the robot has to walk to a predetermined position;
    \end{enumerate}
    \item \textit{Quadcopter (Crazyflie~\cite{crazyflie})}:
    \begin{enumerate}
        \setcounter{enumi}{6}
        \item \textit{Hovering}: the quadcopter should fly to and hover at a designated position (Fig.~\ref{fig.crazyflie});
        \item \textit{Flying through a wind field}: the quadcopter needs to reach a target while flying through a wind field;
        \item \textit{Velocity tracking}: the quadcopter should maintain a specified velocity during flight; 
    \end{enumerate}
\end{itemize}

For each task, we compare the reward functions obtained by using three different methods:
1) $R_{\mathrm{Initial}}$, which is the LLM's initial design of the reward function based on natural language input;
2) $R_{\mathrm{Refined}}$, which is the final reward function formulated by the proposed self-refined LLM framework;
3) $R_{\mathrm{Manual}}$, which is a manually designed reward function sourced from existing literature or benchmarks. 

In the evaluation process, we utilize the Proximal Policy Optimization (PPO)~\cite{schulman2017proximal} as the DRL algorithm to find the optimal policy $\hat{\pi}_R$ for each reward function. 
To concentrate on analyzing the reward function, we use the same learning parameters and neural network architectures in the training processes for all three reward functions.
These parameters are derived by fine-tuning based on the manually designed reward function to achieve its optimal performance.
For each trained policy $\hat{\pi}_R$, we sample $n_t = 100$ trajectories and compute the corresponding objective metrics and success rates.
The success rate threshold for the overall assessment is set at $95\%$, and the maximum number of self-refinement iterations is selected as 5.

We simulate the robotic control tasks with NVIDIA Isaac Sim~\cite{isaacsim, oige} and employ GPT4 as the underlying LLM.
All experiments are conducted on a laptop equipped with an Intel$^\text{\textregistered}$ Core$^\text{\texttrademark}$ i7-10870H CPU and an NVIDIA RTX 3080 Max-Q GPU with 16 GB VRAM.
Further details regarding the experimental setup are given in Appendix~\ref{sec.appendix_experimental_details}.

\begin{figure*}[t!]
    \centering
    \includegraphics[width=\linewidth]{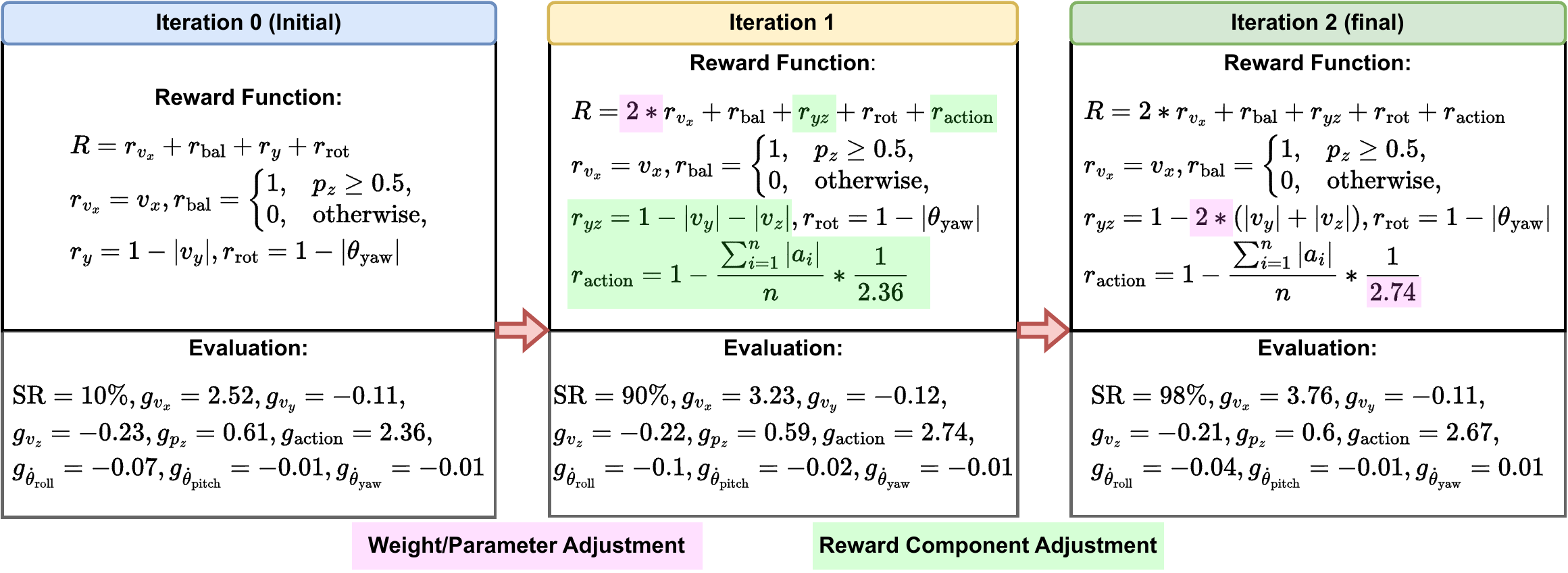}
    \caption{Reward functions in different self-refinement iterations for the quadruped robot forward running task.} 
    \vspace{-10pt}
    \label{fig.reward}
\end{figure*}

\begin{figure*}[t!]
    \centering
    \includegraphics[width=\linewidth]{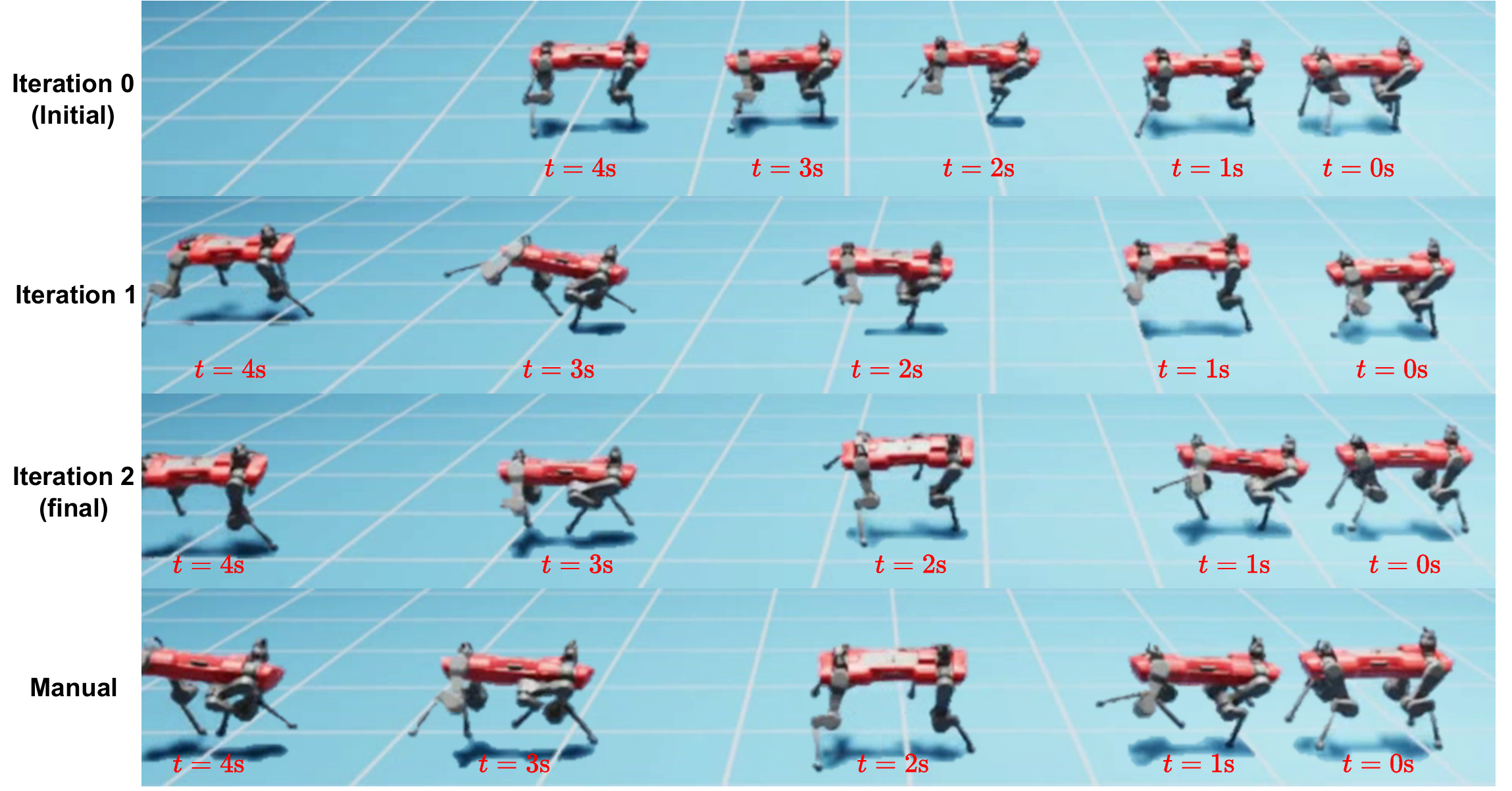}
    \caption{System behaviors corresponding to reward functions in different self-refinement iterations, as well as the manually designed reward function. 
    The time interval between each displayed point is set to 1s.} 
    \label{fig.behaviors}
    \vspace{-10pt}
\end{figure*}

\subsection{Reward Function and Objective Metrics}

We use the quadruped robot forward running task as an example to illustrate the reward function design process via our proposed self-refined LLM framework.
The observable states for this task are:
the global positions of the robot's base $p_x, p_y, p_z$; the linear velocities of the robot $v_x, v_y, v_z$; the base rotations relative to the world frame $\theta_{\mathrm{roll}}, \theta_{\mathrm{pitch}}, \theta_{\mathrm{yaw}}$, the angular velocities $\dot{\theta}_{\mathrm{roll}}, \dot{\theta}_{\mathrm{pitch}}, \dot{\theta}_{\mathrm{yaw}}$, and the current action command for the 12 joints $a_i, i=1\ldots,12$.
The LLM needs to determine which of these observable states should be incorporated into the reward function.
The STL expression representing the success condition is given as 
\begin{equation}
\varphi \equiv \square_{[0.8, 5]} (v_{x} \geq 2) \wedge 	\square_{[0,5]}((p_y \leq 2) \wedge (p_z \geq 0.5)),
\end{equation}
which indicates that following an initial acceleration phase lasting 
0.8 seconds, the robot must always maintain a speed of at least $v_x=2$ m/s until the simulation stops at $t=5$ seconds. 
Meanwhile, the robot must restrict lateral deviation to under $2$ m and cannot fall over.
The objective metrics used in the evaluation process are:
the average linear velocities $g_{v_x}, g_{v_y},g_{v_z}$; the average $z$-position $g_{p_z}$; the average normalized action values $g_{\mathrm{action}}$; and the average angular velocities $g_{\dot{\theta}_{\mathrm{roll}}}, g_{\dot{\theta}_{\mathrm{pitch}}}, g_{\dot{\theta}_{\mathrm{yaw}}}$.
For comparison, we employ a manually designed reward function based on~\cite{oige}, which is given as
\begin{equation}
    R = 1.5v_x + 0.2r_{\mathrm{bal}} - 0.5\frac{|p_y|}{2} - 0.1|\dot{\theta}_{\mathrm{yaw}}|,
\end{equation}
with $r_{\mathrm{bal}} = 1$ if $p_z \geq 0.5$ and $r_{\mathrm{bal}} = 0$ otherwise.

The LLM requires a total of two self-refinement iterations to identify a satisfactory reward function. 
Fig.~\ref{fig.reward} presents the reward functions in different iterations alongside their respective evaluation outcomes. 
Corresponding system behaviors are given in Fig.~\ref{fig.behaviors}, which also illustrates the behavior produced by the manually designed reward function.

Similar to a manual refinement process, the LLM adjusts the reward function by altering its weights or parameters or by modifying the structure of its components. 
Initially, the reward function contains four components, emphasizing forward velocity and balance.
However, this initial design proves insufficient, as the robot overly prioritizes maintaining balance over running forward, leading to a low forward velocity and success rate (see Iteration 0 in Fig.~\ref{fig.behaviors}).
Responding to this feedback, our proposed framework then initiates a self-refinement iteration. 
It increases the weight attributed to forward velocity $v_x$ and adjusts the penalty associated with lateral deviation. 
Furthermore, it introduces a penalty for large actions in reaction to the action metric $g_{\mathrm{action}}$ contained in the feedback. 
This refinement enhances performance, elevating the success rate to 90\%.
However, the evaluation indicates that the robot still takes aggressive actions to achieve high velocity (e.g., $t=3$s of Iteration 1 in Fig.~\ref{fig.behaviors}).
Hence, in its second self-refinement iteration, the LLM increases the penalties for excessive actions and lateral deviations. 
The resulting reward function leads to a behavior that closely aligns with the behavior obtained from a manually designed reward function (see Iteration 2 and Manual in Fig.~\ref{fig.behaviors}).
The manually designed reward function yields the following evaluation metrics: $ \mathrm{SR}=95\% ,g_{v_x} = 3.748, g_{v_y} = -0.105, g_{v_z} = -0.223, g_{p_z} = 0.609, g_{\mathrm{action}} = 2.673, g_{\dot{\theta}_{\mathrm{roll}}} = -0.071, g_{\dot{\theta}_{\mathrm{pitch}}} =-0.019, g_{\dot{\theta}_{\mathrm{yaw}}} =0.041$, which are also similar to those of the final reward function formulated by the LLM.
This justifies the efficacy of the proposed self-refined LLM framework in designing reward functions for continuous robotic control tasks.

Other control tasks exhibit a similar pattern as the quadruped robot forward running task.
See Appendix~\ref{sec.appendix_experimental_log} and the supplementary video for details on these tasks.

\subsection{Success Rates}
\label{sec.success_rate}

\begin{table}[t]
\centering
\caption{Success rates of different reward functions and the number of self-refinement iterations (Iter.) used for $R_{\mathrm{Refined}}$.}
\label{table.results}
\resizebox{.9\linewidth}{!}{
\begin{tabular}{cc|ccc|c}
\toprule
& &   \multicolumn{3}{c|}{Success Rate $\mathrm{SR}$} & \\
Robotic System & Task & $R_{\mathrm{Initial}}$ & $R_{\mathrm{Refined}}$ & $R_{\mathrm{Manual}}$ & Iter. \\ \midrule
\multirow{3}{*}{Manipulator} & Ball Catching & 100$\%$ & 100$\%$ & 100$\%$ & 0 \\  
& Ball Balancing & 100$\%$ & 100$\%$ & 98$\%$ & 0 \\
& Ball Pushing & 0$\%$ & 93$\%$ & 95$\%$ & 5\\ \midrule
\multirow{4}{*}{Quadruped} & Velocity Tracking  & 0$\%$ & 96$\%$ & 92$\%$ & 3 \\  
& Running  &  10$\%$ &  98$\%$ & 95$\%$  & 2\\
& Walking to Target  &  0$\%$ &  85$\%$ & 80$\%$ & 5\\ \midrule
\multirow{3}{*}{Quadcopter} & Hovering  &  0$\%$ &  98$\%$ & 92$\%$ & 2\\  
& Wind Field  &  0$\%$ &  100$\%$ & 100$\%$ & 4\\
& Velocity Tracking  & 0$\%$ &  99$\%$ & 91$\%$ & 3\\ \bottomrule
\end{tabular}
}
\vspace{-10pt}
\end{table}

We further evaluate the success rates across all the tasks under consideration to assess the generalizability and applicability of our proposed approach.
The results are summarized in Table~\ref{table.results}. 
Detailed information regarding the employed success conditions for each task is presented in Appendix~\ref{sec.appendix_experimental_details}.

It can be observed that the initial reward function demonstrates a binary level of performance. 
For tasks with straightforward objectives, such as ball catching or balancing, the LLM is able to devise a high-performing reward function on its first attempt.
Conversely, for more complex tasks that involve multiple objectives, e.g., ensuring a quadruped robot maintains a set velocity while walking straight and keeping balance, the initial reward function often registers a success rate of $0\%$. 
In such cases, the LLM predominantly relies on feedback to understand the implications of its design, necessitating multiple self-refinement iterations. 
By leveraging the evaluation results, the LLM is capable of effectively revising its reward function design.
As a result, it achieves success rates that match or even surpass those of manually designed reward functions for all examined tasks.

However, the performance is affected by the intrinsic complexities of the task.
For tasks demanding intricate reward function components, e.g., the quadruped robot walking to target task, the success rate diminishes, indicating a need for further self-refinement iterations or even detailed human feedback.
Nevertheless, even in these challenging scenarios, our self-refined LLM framework consistently identifies reward functions that outperform manual designs.
This illustrates the broad applicability of our approach across a wide range of continuous robotic control tasks.

\section{Discussion}
\label{sec.discussion}
\noindent
\textbf{Learning Parameters and AutoRL}
In our experiments, we observe that during the self-refinement process, the LLM often has to adjust the weights of reward components. 
While the LLM is able to determine a final reward function that yields desired system behavior, the weights it assigns are not guaranteed to be optimal. 
In other words, there might exist configurations that produce even better outcomes. 
One potential improvement would be to integrate the LLM with AutoRL. 
Once the LLM formulates the reward function, AutoRL could optimize its parameters using search-based approaches. 
In such a case, the LLM serves as an initial designer, offering a parameterized reward function to AutoRL. 
This strategy can further be extended to fine-tune learning parameters and neural network architectures. 
By identifying optimal parameters before each self-refinement iteration, the LLM can then focus on adjusting the structural components of the reward function. 
However, adopting this approach could greatly prolong the reward function design process.

\noindent
\textbf{Fine-tuned LLM} 
Recent studies indicate that fine-tuning the LLM for specific tasks can greatly enhance its performance~\cite{houlsby2019parameter,li2021prefix,ouyang2022training}. 
Such a technique could also improve our approach. 
The complexity associated with comprehending control task requirements and formulating appropriate reward functions could potentially be alleviated by deploying an LLM specifically fine-tuned for reward function design, as opposed to a general-purpose model. 
Nonetheless, fine-tuning an LLM typically demands substantial resources, and garnering enough training data for reward function design might also be challenging.

\noindent
\textbf{Limitations}
One major limitation of our approach is its inability to address nuanced aspects of desired system behaviors that are difficult to quantify through the automated evaluation process, such as the gait of a quadruped robot.
Addressing this challenge often necessitates human intervention.
By offering detailed human feedback, the LLM is capable of fine-tuning its outcome accordingly, as illustrated in~\cite{yu2023language}.
Another limitation is the reliance of the LLM on its pre-trained common-sense knowledge. 
For tasks that are highly specialized or not represented in its training data, the LLM may struggle to devise an appropriate reward function.
Under such circumstances, enhancing the natural language input prompt with more details about the specific robotic system and control task becomes essential.

\section{Conclusion}
\label{sec.conclusion}

In this paper, we introduce a self-refined LLM framework as an automated reward function designer for DRL in continuous robotic control tasks.
The framework operates in three steps:
First, the LLM devises an initial reward function by using a natural language input.
Second, an automated evaluation process is initiated to assess the performance of the designed reward function.
Third, based on the evaluation results, a feedback prompt is provided to the LLM, guiding its self-refinement process of the reward function.
We evaluate our proposed framework across nine diverse robotic control tasks, distributed among three distinct robotic systems.
The results indicate that our approach is able to generate reward functions that are on par with, or even superior to, those manually designed ones.
For future work, we plan to integrate the LLM with AutoRL techniques, enabling not only the reward function, but also all learning parameters to be designed autonomously.

\bibliographystyle{ACM-Reference-Format}
\bibliography{ref}

\appendix
\section{Natural Language Input Prompt}
\label{sec.appendix_input_prompt}

The natural language input prompts used for each control task are given as follows.

\subsection{Robotic Manipulator}

\begin{tcolorbox}[title={Ball Catching},breakable] \label{ball_catching_prompt}
I want to design a reward function for a reinforcement learning task. \\

I have a 7 DOF manipulator with a gripper attached as the end effector. The manipulator initially holds a container that opens upwards, and a ball will be thrown from above. \\

This task has the following goals:

1. The manipulator should use the container to capture the ball before it falls on the ground. \\
2. When the ball has been captured, it should stay within the container as long as possible. \\

The following variables are available for reward function design.

1. ball\_pos: This vector with a dimension of 3 represents the x,y,z position of the ball according to the world coordinate. 

2. ball\_vel: This vector contains the linear velocity of the ball according to the world coordinate. 

3. container\_pos: This vector represents the x,y,z position of the container according to the world coordinate.

4. container\_rot: This 4-dimensional vector represents the rotation of the container in quaternion. \\

Some rules while designing the reward function: 

1) The reward function should only use the given variables.

2) The reward function should not use unknown quantities, such as desired joint positions, in its computation.\\

Design a complete reward function for this task.
\end{tcolorbox}


\begin{tcolorbox}[title={Ball Balancing},breakable]  \label{ball_balancing_prompt}
I want to design a reward function for a reinforcement learning task. \\

I have a 7 DOF manipulator with a gripper attached as the end effector. The manipulator initially holds a tray, and a ball will fall from above. \\

This task has the following goals:\\
1. The manipulator should use the tray to catch the ball and balance it at the center of the tray.

2. When the ball has been caught, it should be held stably in the center of the tray as long as possible. \\

The following variables are available for reward function design.

1. ball\_pos: This vector with a dimension of 3 represents the x,y,z position of the ball according to the world coordinate. 

2. ball\_vel: This vector contains the linear velocity of the ball according to the world coordinate. 

3. tray\_pos: This vector represents the x,y,z position of the tray according to the world coordinate.

4. tray\_rot: This 4-dimensional vector represents the rotation of the tray in quaternion. \\

Some rules while designing the reward function: 

1) The reward function should only use the given variables.

2) The reward function should not use unknown quantities, such as desired joint positions, in its computation.\\

Design a complete reward function for this task.
\end{tcolorbox}


\begin{tcolorbox}[title={Ball Pushing},breakable] \label{ball_pushing_prompt}
I want to design a reward function for a reinforcement learning task. \\

I have a 7 DOF manipulator with a gripper attached as the end effector. A table is placed on the ground in front of the manipulator.
A ball is placed on the table. A hole exists in the tale. \\

This task has the following goals:\\
1. The manipulator is required to push the ball to make it drop into the hole in the table.\\

The following variables are available for reward function design.

1. gripper\_pos: This vector with a dimension of 3 represents the x,y,z position of the gripper according to the world coordinate. 

2. ball\_pos: This vector with a dimension of 3 represents the x,y,z position of the ball according to the world coordinate. 

3. hole\_pos: This vector with a dimension of 3 represents the x,y,z position of the target hole according to the world coordinate. 

4. ball\_vel: This vector contains the linear velocity of the ball according to the world coordinate. \\

Some rules while designing the reward function: 

1) The reward function should only use the given variables.

2) The reward function should not use unknown quantities, such as desired joint positions, in its computation.\\

Design a complete reward function for this task.
\end{tcolorbox}

\subsection{Quadruped Robot}


\begin{tcolorbox}[title={Velocity Tracking},breakable] \label{anymal_velocity_prompt}
I want to design a reward function for a reinforcement learning task. \\

I have a quadruped robot.  A target linear velocity is given among the x, y and z axes.\\

This task has the following goals: \\
1. The quadruped robot should track the specified linear velocity.

2. The quadruped robot cannot fall over (the position of the robot on the z-axis cannot be lower than 0.5).

3. The quadruped robot should walk stably and smoothly (the robot should not jump or rotate in the air) \\

The following variables are available for reward function design.

1. robot\_pos: This vector with a dimension of 3 represents the x,y,z position of the quadruped robot according to the world coordinate.  

2. robot\_rot: This 4-dimensional vector represents the rotation of the quadruped robot in quaternion.

3. robot\_linvel: This 3-dimensional vector represents the linear velocity of the quadruped robot in x, y and z axes.

4. robot\_angvel: This 3-dimensional vector represents the angular velocity of the quadruped robot in roll, yaw and pitch.

5. actions: This 12-dimensional vector represents the action output of the RL agent, which are the 12 joint positions of the quadruped robot.

6. target\_vel: This vector with a dimension of 3 represents the target linear velocity that the Quadcopter needs to track along x, y and z axes \\

Some rules while designing the reward function: 

1) The reward function should only use the given variables.

2) The reward function should not use unknown quantities, such as desired joint positions, in its computation.\\

Design a complete reward function for this task.
\end{tcolorbox}


\begin{tcolorbox}[title={Running},breakable] \label{anymal_running_prompt}
I want to design a reward function for a reinforcement learning task. \\

I have a quadruped robot. \\

This task has the following goals: \\
1. The quadruped robot should run forward in a straight line along the x-axis as fast as possible.

2. The quadruped robot cannot fall over (the position of the robot on the z-axis cannot be lower than 0.5).

3. The quadruped robot should avoid large deviations along the y-axis. \\

The following variables are available for reward function design.

1. robot\_pos: This vector with a dimension of 3 represents the x,y,z position of the quadruped robot according to the world coordinate.  

2. robot\_rot: This 4-dimensional vector represents the rotation of the quadruped robot in quaternion.

3. robot\_linvel: This 3-dimensional vector represents the linear velocity of the quadruped robot in the x, y and z axes.

4. robot\_angvel: This 3-dimensional vector represents the angular velocity of the quadruped robot in roll, yaw and pitch.

5. actions: This 12-dimensional vector represents the action output of the RL agent, which are the 12 joint positions of the quadruped robot.\\

Some rules while designing the reward function: 

1) The reward function should only use the given variables.

2) The reward function should not use unknown quantities, such as desired joint positions, in its computation.\\

Design a complete reward function for this task.
\end{tcolorbox}


\begin{tcolorbox}[title={Walking to Target},breakable] \label{anymal_walkingg_to_position_prompt}
I want to design a reward function for a reinforcement learning task. \\

I have a quadruped robot. A target position is given in 3D Cartesian space. \\

This task has the following goals: \\
1. The quadruped robot should walk to the target position and stay at that position (The normalized distance on the x and y axes between the robot and the target position should be less than 0.2).

2. The quadruped robot cannot fall over (the position of the robot on the z-axis cannot be lower than 0.5).\\

The following variables are available for reward function design.

1. robot\_pos: This vector with a dimension of 3 represents the x,y,z position of the quadruped robot according to the world coordinate.  

2. robot\_rot: This 4-dimensional vector represents the rotation of the quadruped robot in quaternion.

3. robot\_linvel: This 3-dimensional vector represents the linear velocity of the quadruped robot in the x, y and z axes.

4. robot\_angvel: This 3-dimensional vector represents the angular velocity of the quadruped robot in roll, yaw and pitch.

5. actions: This 12-dimensional vector represents the action output of the RL agent, which are the 12 joint positions of the quadruped robot.

6. target\_pos: This vector with a dimension of 3 represents the x,y,z position of the target position according to the world coordinate. \\

Some rules while designing the reward function: 

1) The reward function should only use the given variables.

2) The reward function should not use unknown quantities, such as desired joint positions, in its computation.

3) The reward should only contain positive components. \\

Design a complete reward function for this task.
\end{tcolorbox}

\subsection{Quadcopter}

\begin{tcolorbox}[title={Hovering},breakable] \label{crazyflie_hovering_prompt}
I want to design a reward function for a reinforcement learning task. \\

I have a Quadcopter robot. A target position is given in 3D Cartesian space.  \\

This task has the following goals: \\
1. The Quadcopter should fly to the target position then hover at that position.

2. The Quadcopter cannot fall on the ground and cannot fly too high. (The robot position on the z-axis should be less than 3.0 and greater than 0.8.)\\

The following variables are available for reward function design.

1. copter\_pos: This vector with a dimension of 3 represents the x,y,z position of the Quadcopter according to the world coordinate. 

2. copter\_rot: This 4-dimensional vector represents the rotation of the Quadcopter in quaternion.

3. target\_pos: This vector with a dimension of 3 represents the x,y,z position of the target position according to the world coordinate. 

4. copter\_angvels: This 4-dimensional vector represents the angular velocity of the Quadcopter in quaternion.

5. actions: This vector represents the action output of the RL agent which is the thrust forces to the four rotors of the Quadcopter.

6. copter\_linvels: This 3-dimensional vector represents the linear velocity of the Quadcopter in x, y and z axes.\\

Some rules while designing the reward function: 

1) The reward function should only use the given variables.

2) The reward function should not use unknown quantities, such as desired joint positions, in its computation.\\

Design a complete reward function for this task.
\end{tcolorbox}


\begin{tcolorbox}[title={Wind Field},breakable] \label{crazyflie_wind_prompt}
I want to design a reward function for a reinforcement learning task. \\

I have a Quadcopter. A target position is given in 3D Cartesian space. Wind disturbance exists within an area in front of the Quadcopter. The Quadcopter has to fly through this area to reach the target position.\\

This task has the following goals: \\
1. The manipulator is required to push the ball to make it drop into the hole in the table.\\

The following variables are available for reward function design.

1. copter\_pos: This vector with a dimension of 3 represents the x,y,z position of the Quadcopter according to the world coordinate. 

2. copter\_rot: This 4-dimensional vector represents the rotation of the Quadcopter in quaternion.

3. target\_pos: This vector with a dimension of 3 represents the x,y,z position of the target position according to the world coordinate. 

4. copter\_angvels: This 4-dimensional vector represents the angular velocity of the Quadcopter in quaternion.

5. actions: This vector represents the action output of the RL agent which is the thrust forces to the four rotors of the Quadcopter.

6. copter\_linvels: This 3-dimensional vector represents the linear velocity of the Quadcopter in x, y and z axes.\\ \\

Some rules while designing the reward function: 

1) The reward function should only use the given variables.

2) The reward function should not use unknown quantities, such as desired joint positions, in its computation.\\

Design a complete reward function for this task.
\end{tcolorbox}


\begin{tcolorbox}[title={Velocity Tracking},breakable] \label{crazyflie_velocity_prompt}
I want to design a reward function for a reinforcement learning task. \\

I have a Quadcopter. A specified Quadcopter linear velocity is given among the x, y and z axes. \\

This task has the following goals: \\
1. The Quadcopter should track the target linear velocity.

2. The Quadcopter should maintain a constant height (The position of the Quadcopter on the z-axis should be fixed to 1.0)\\

The following variables are available for reward function design.

1. copter\_pos: This vector with a dimension of 3 represents the x,y,z position of the Quadcopter according to the world coordinate.  
2. copter\_rot: This 4-dimensional vector represents the rotation of the Quadcopter in quaternion.

3. target\_vel: This vector with a dimension of 3 represents the target linear velocity that the Quadcopter needs to track along x, y and z axes 

4. copter\_angvels: This 4-dimensional vector represents the angular velocity of the Quadcopter in quaternion.

5. actions: This vector represents the action output of the RL agent, which is the thrust forces to the four rotors of the Quadcopter.

6. copter\_linvels: This 3-dimensional vector represents the linear velocity of the Quadcopter in x, y and z axes.\\

Some rules while designing the reward function: 

1) The reward function should only use the given variables.

2) The reward function should not use unknown quantities, such as desired joint positions, in its computation.\\

Design a complete reward function for this task.
\end{tcolorbox}


\section{Feedback Prompt Template}
\label{sec.appendix_feedback_prompt}

The feedback prompt templates used for each control task are given as follows.

\subsection{Robotic Manipulator}
\begin{tcolorbox}[title={Ball Catching},breakable] \label{ball_catching_feedback}
The performance of the RL agent trained with the designed reward function is [\textit{good|bad}].
\\

The training reward converges after [\textit{NUM}] steps. \\
The average timestep for one episode is [\textit{NUM}]. \\
The average reward is [\textit{NUM}] for one episode. \\

The average normalized action value is [\textit{NUM}] \\
The normalized distance between the ball and the container is [\textit{NUM}] \\
The average distance between the ball and the container on the x-axis is [\textit{NUM}] \\
The average distance between the ball and the container on the y-axis is [\textit{NUM}] \\
The average distance between the ball and the container on the z-axis is [\textit{NUM}] \\
The ball's average linear velocity on the x-axis is [\textit{NUM}] \\
The ball's average linear velocity on the y-axis is [\textit{NUM}] \\
The ball's average linear velocity on the z-axis is [\textit{NUM}] \\

Evaluation results of [\textit{NUM}] trials are given below: \\
Goal 1 success rate is [\textit{NUM}] \\

Redesign the reward function based on the given feedback.
\end{tcolorbox}

\begin{tcolorbox}[title={Ball Balancing},breakable] \label{ball_balancing_feedback}
The performance of the RL agent trained with the designed reward function is [\textit{good|bad}]. \\

The training reward converges after [\textit{NUM}] steps. \\
The average timestep for one episode is [\textit{NUM}]. \\
The average reward is [\textit{NUM}] for one episode. \\

The average normalized action value is [\textit{NUM}] \\
The normalized distance between the ball and the center of the tray is [\textit{NUM}] \\
The average distance between the ball and the center of the tray on the x-axis is [\textit{NUM}] \\
The average distance between the ball and the center of the tray on the y-axis is [\textit{NUM}] \\
The average distance between the ball and the center of the tray on the z-axis is [\textit{NUM}] \\
The ball's average linear velocity on the x-axis is [\textit{NUM}] \\
The ball's average linear velocity on the y-axis is [\textit{NUM}] \\
The ball's average linear velocity on the z-axis is [\textit{NUM}] \\

Evaluation results of [\textit{NUM}] trials are given below: \\
Goal 1 success rate is [\textit{NUM}] \\
Goal 2 success rate is [\textit{NUM}] \\

Redesign the reward function based on the given feedback.
\end{tcolorbox}

\begin{tcolorbox}[title={Ball Pushing},breakable] \label{ball_pushing_feedback}
The performance of the RL agent trained with the designed reward function is [\textit{good|bad}]. \\

The training reward converges after [\textit{NUM}] steps. \\
The average timestep for one episode is [\textit{NUM}]. \\
The average reward is [\textit{NUM}] for one episode. \\

The average normalized action value is [\textit{NUM}] \\
The normalized distance between the ball and the target hole is [\textit{NUM}] \\
The average distance between the ball and the target hole on the x-axis is [\textit{NUM}] \\
The average distance between the ball and the target hole on the y-axis is [\textit{NUM}] \\
The average distance between the ball and the target hole on the z-axis is [\textit{NUM}] \\
The ball's average linear velocity on the x-axis is [\textit{NUM}] \\
The ball's average linear velocity on the y-axis is [\textit{NUM}] \\
The ball's average linear velocity on the z-axis is [\textit{NUM}] \\

Evaluation results of [\textit{NUM}] trials are given below: \\
Goal 1 success rate is [\textit{NUM}] \\
Goal 2 success rate is [\textit{NUM}] \\

Redesign the reward function based on the given feedback.
\end{tcolorbox}

\subsection{Quadruped Robot}

\begin{tcolorbox}[title={Velocity Tracking},breakable] \label{anymal_velocity_feedback}
The performance of the RL agent trained with the designed reward function is [\textit{good|bad}]. \\

The training reward converges after [\textit{NUM}] steps. \\
The average timestep for one episode is [\textit{NUM}]. \\
The average reward is [\textit{NUM}] for one episode. \\

The normalized linear velocity deviation is [\textit{NUM}] \\
The average normalized action value is [\textit{NUM}] \\
The robot's average linear velocity on the x-axis is [\textit{NUM}] \\
The robot's average linear velocity on the y-axis is [\textit{NUM}] \\
The robot's average linear velocity on the z-axis is [\textit{NUM}] \\
The robot's average angular velocity on the x-axis is [\textit{NUM}] \\
The robot's average angular velocity on the y-axis is [\textit{NUM}] \\
The robot's average angular velocity on the z-axis is [\textit{NUM}] \\
The robot's average position on the z-axis is [\textit{NUM}] \\

Evaluation results of [\textit{NUM}] trials are given below: \\
Goal 1 success rate is [\textit{NUM}] \\
Goal 2 success rate is [\textit{NUM}] \\
Goal 3 success rate is [\textit{NUM}] \\

Redesign the reward function based on the given feedback.
\end{tcolorbox}

\begin{tcolorbox}[title={Running},breakable] \label{anymal_running_feedback}
The performance of the RL agent trained with the designed reward function is [\textit{good|bad}]. \\

The training reward converges after [\textit{NUM}] steps. \\
The average timestep for one episode is [\textit{NUM}]. \\
The average reward is [\textit{NUM}] for one episode. \\

The robot's maximum linear velocity along the x-axis is [\textit{NUM}] \\
The average normalized action value is [\textit{NUM}] \\
The robot's average linear velocity on the x-axis is [\textit{NUM}] \\
The robot's average linear velocity on the y-axis is [\textit{NUM}] \\
The robot's average linear velocity on the z-axis is [\textit{NUM}] \\
The robot's average angular velocity on the x-axis is [\textit{NUM}] \\
The robot's average angular velocity on the y-axis is [\textit{NUM}] \\
The robot's average angular velocity on the z-axis is [\textit{NUM}] \\
The robot's average position on the z-axis is [\textit{NUM}] \\

Evaluation results of [\textit{NUM}] trials are given below: \\
Goal 1 success rate is [\textit{NUM}] \\
Goal 2 success rate is [\textit{NUM}] \\
Goal 3 success rate is [\textit{NUM}] \\

Redesign the reward function based on the given feedback.
\end{tcolorbox}


\begin{tcolorbox}[title={Walking to Target},breakable] \label{anymal_walking_to_position_feedback}
The performance of the RL agent trained with the designed reward function is [\textit{good|bad}]. \\

The training reward converges after [\textit{NUM}] steps. \\
The average timestep for one episode is [\textit{NUM}]. \\
The average reward is [\textit{NUM}] for one episode. \\

The normalized distance between the robot and the target position is [\textit{NUM}] \\
The average normalized action value is [\textit{NUM}] \\
The robot's average linear velocity on the x-axis is [\textit{NUM}] \\
The robot's average linear velocity on the y-axis is [\textit{NUM}] \\
The robot's average linear velocity on the z-axis is [\textit{NUM}] \\
The robot's average angular velocity on the x-axis is [\textit{NUM}] \\
The robot's average angular velocity on the y-axis is [\textit{NUM}] \\
The robot's average angular velocity on the z-axis is [\textit{NUM}] \\
The robot's average position on the z-axis is [\textit{NUM}] \\

Evaluation results of [\textit{NUM}] trials are given below: \\
Goal 1 success rate is [\textit{NUM}] \\
Goal 2 success rate is [\textit{NUM}] \\

Redesign the reward function based on the given feedback.
\end{tcolorbox}

\subsection{Quadcopter}
\begin{tcolorbox}[title={Hovering},breakable] \label{crazyflie_hovering_feedback} 
The performance of the RL agent trained with the designed reward function is [\textit{good|bad}]. \\

The training reward converges after [\textit{NUM}] steps. \\
The average timestep for one episode is [\textit{NUM}]. \\
The average reward is [\textit{NUM}] for one episode. \\

The normalized distance between the robot and the target position is [\textit{NUM}] \\
The average normalized action value is [\textit{NUM}] \\
The robot's average linear velocity on the x-axis is [\textit{NUM}] \\
The robot's average linear velocity on the y-axis is [\textit{NUM}] \\
The robot's average linear velocity on the z-axis is [\textit{NUM}] \\
The robot's average angular velocity on the x-axis is [\textit{NUM}] \\
The robot's average angular velocity on the y-axis is [\textit{NUM}] \\
The robot's average angular velocity on the z-axis is [\textit{NUM}] \\
The robot's average position on the z-axis is [\textit{NUM}] \\

Evaluation results of [\textit{NUM}] trials are given below: \\
Goal 1 success rate is [\textit{NUM}] \\
Goal 2 success rate is [\textit{NUM}] \\

Redesign the reward function based on the given feedback.
\end{tcolorbox}

\begin{tcolorbox}[title={Wind Field},breakable] \label{crazyflie_wind_feedback}
The performance of the RL agent trained with the designed reward function is [\textit{good|bad}]. \\

The training reward converges after [\textit{NUM}] steps. \\
The average timestep for one episode is [\textit{NUM}]. \\
The average reward is [\textit{NUM}] for one episode. \\

The normalized distance between the robot and the target position is [\textit{NUM}] \\
The average normalized action value is [\textit{NUM}] \\
The robot's average linear velocity on the x-axis is [\textit{NUM}] \\
The robot's average linear velocity on the y-axis is [\textit{NUM}] \\
The robot's average linear velocity on the z-axis is [\textit{NUM}] \\
The robot's average angular velocity on the x-axis is [\textit{NUM}] \\
The robot's average angular velocity on the y-axis is [\textit{NUM}] \\
The robot's average angular velocity on the z-axis is [\textit{NUM}] \\
The robot's average position on the z-axis is [\textit{NUM}] \\

Evaluation results of [\textit{NUM}] trials are given below: \\
Goal 1 success rate is [\textit{NUM}] \\
Goal 2 success rate is [\textit{NUM}] \\

Redesign the reward function based on the given feedback.
\end{tcolorbox}

\begin{tcolorbox}[title={Velocity Tracking},breakable] \label{crazyflie_velocity_feedback}
The performance of the RL agent trained with the designed reward function is [\textit{good|bad}]. \\

The training reward converges after [\textit{NUM}] steps. \\
The average timestep for one episode is [\textit{NUM}]. \\
The average reward is [\textit{NUM}] for one episode. \\

The average normalized action value is [\textit{NUM}] \\
The normalized linear velocity deviation is [\textit{NUM}] \\
The linear velocity deviation on the x-axis is [\textit{NUM}] \\
The linear velocity deviation on the y-axis is [\textit{NUM}] \\
The linear velocity deviation on the z-axis is [\textit{NUM}] \\
The robot's average angular velocity on the x-axis is [\textit{NUM}] \\
The robot's average angular velocity on the y-axis is [\textit{NUM}] \\
The robot's average angular velocity on the z-axis is [\textit{NUM}] \\
The robot's average position on the z-axis is [\textit{NUM}] \\

Evaluation results of [\textit{NUM}] trials are given below: \\
Goal 1 success rate is [\textit{NUM}] \\
Goal 2 success rate is [\textit{NUM}] \\

Redesign the reward function based on the given feedback.
\end{tcolorbox}


\section{Experimental Details}
\label{sec.appendix_experimental_details}

In this section, we explain the experimental details associated with each control task.

\subsection{Robotic Manipulator}

\begin{tcolorbox}[title={Ball Catching},breakable]
\textbf{Environmental Setup}

The base of the manipulator is located at (0,0,0) in world coordinates. The manipulator initially holds a container that opens upwards, and a ball will be thrown from above.

The initial position of the ball is randomly located within an area of ($x = [-0.05, 0.05], y = [-0.05, 0.05], z=1.5$).

The ball has a randomized initial linear velocity of $(v_x = [0.5, 1.0], v_y = 0.0, 0.2, v_z = 0)$.

The observable states for this task can be found in the natural language input to the LLM (see Appendix ~\ref{ball_catching_prompt}).

The objective metrics for evaluation are defined in the feedback prompt presented in Appendix ~\ref{ball_catching_feedback}, \\

\textbf{Task Specification (STL)}

The success condition for this task is defined as:
\begin{align*}
    & \varphi_1 \equiv \square_{[3, 30]} (\parallel \text{ball}_{pos} - \text{tool}_{pos} \parallel \leq 0.1) \\
    & \varphi_2 \equiv \square_{[0, 30]} (\text{ball}_{pos-z} \geq 0.2) \\
    & \varphi_{SR} \equiv \varphi_2 \wedge \varphi_2 
\end{align*}
where $\varphi_1$ means that following an initial catching phase of 3 seconds, the distance between the ball and the center of the bottom of the container should be less than 0.1 m until the simulation stops at 30 seconds, $\varphi_2$ means the ball cannot fall on the ground during the entire simulation, and $\varphi_{SR}$ means $\varphi_1$ and $\varphi_2$ must be held simultaneously to indicate the task completes successfully.  
 
\end{tcolorbox}

\begin{tcolorbox}[title={Ball Balancing},breakable]

\textbf{Environmental Setup}

The base of the manipulator is located at (0,0,0) in world coordinates.   The manipulator initially holds a tray, and a ball will fall from above.

The initial position of the ball is randomly located within an area of ($x = [0.2, 0.5], y = [-0.15, 0.15], z=1.5$).

The observable states for this task can be found in the natural language input to the LLM (see Appendix ~\ref{ball_balancing_prompt}).

The objective metrics for evaluation are defined in the feedback prompt presented in Appendix ~\ref{ball_balancing_feedback}, \\

\textbf{Task Specification (STL)}

The success condition for this task is defined as:
\begin{align*}
    & \varphi_1 \equiv \square_{[3, 30]} (\parallel \text{ball}_{pos} - \text{tray}_{pos} \parallel \leq 0.25) \\
    & \varphi_2 \equiv \square_{[0, 30]} (\text{ball}_{pos-z} \geq 0.2) \\
    & \varphi_{SR} \equiv \varphi_2 \wedge \varphi_2 
\end{align*}
where $\varphi_1$ means that following an initial phase of 3 seconds, the distance between the ball and the center of the tray should be less than 0.25 m until the simulation stops at 30 seconds, $\varphi_2$ means the ball cannot fall on the ground during the entire simulation, and $\varphi_{SR}$ means $\varphi_1$ and $\varphi_2$ must be held simultaneously to indicate the task completes successfully. 
\end{tcolorbox}

\begin{tcolorbox}[title={Ball Pushing},breakable]

\textbf{Environmental Setup}

The base of the manipulator is located at (0,0,0) in world coordinates.  A table is placed in front of the manipulator along the x-axis. A table is placed on the ground in front of the manipulator. A ball is placed on the one side of the table. A hole exists on the other side of the table.

The initial position of the ball is randomly located within an area of ($x = [0.4, 0.6], y = [-0.1, 0.11], z = 0.45$). There are no initial linear or angular velocities applied to the ball.

The observable states for this task can be found in the natural language input to the LLM (see Appendix ~\ref{ball_pushing_prompt}).

The objective metrics for evaluation are defined in the feedback prompt presented in Appendix ~\ref{ball_pushing_feedback}, \\

\textbf{Task Specification (STL)}

The success condition for this task is defined as:
\begin{align*}
    & \varphi_1 \equiv \Diamond_{[0, 30]} (\parallel \text{ball}_{pos} - \text{hole}_{pos} \parallel \leq 0.12) \\
    & \varphi_2 \equiv \square_{[0, 30]} (\text{ball}_{pos-z} \geq 0.2) \\
    & \varphi_{SR} \equiv \varphi_2 \wedge \varphi_2 
\end{align*}

where $\varphi_1$ means that eventually, the distance between the ball and the center of the hole should be less than 0.12 m until the simulation stops at 30 seconds, $\varphi_2$ means the ball cannot fall on the ground during the entire simulation, and $\varphi_{SR}$ means $\varphi_1$ and $\varphi_2$ must be held simultaneously to indicate the task completes successfully. 
\end{tcolorbox}

\subsection{Quadruped Robot}

\begin{tcolorbox}[title={Velocity Tracking},breakable]

\textbf{Environmental Setup}

The initial position of the quadruped robot is located at (0,0,0) in world coordinates.  A specified forward direction of the robot is along with the positive direction of the x-axis. A designated velocity is given along this direction.

The observable states for this task can be found in the natural language input to the LLM (see Appendix ~\ref{anymal_velocity_prompt}).

The objective metrics for evaluation are defined in the feedback prompt presented in Appendix ~\ref{anymal_velocity_feedback}, \\

\textbf{Task Specification (STL)}

The success condition for this task is defined as:
\begin{align*}
    & \varphi_1 \equiv \square_{[0.8, 5]} (|\text{robot}_{v-x} - \text{target}_{v}| \leq 0.5) \\
    & \varphi_2 \equiv \square_{[0, 5]} (|\text{robot}_{pos-y}| \leq 2) \\
    & \varphi_3 \equiv \square_{[0, 5]} (\text{robot}_{pos-z} \geq 0.5) \\
    & \varphi_{SR} \equiv \varphi_2 \wedge \varphi_2 \wedge \varphi_3
\end{align*}

where $\varphi_1$ means that following an initial acceleration phase of 0.8 seconds, the velocity deviation between the robot and the specified target velocity on the x-axis should be less than 0.5 m/s until the simulation stops at 5 seconds, $\varphi_2$ means that the lateral deviation of the robot should be less than 2 m throughout the simulation, $\varphi_3$ means the robot cannot fall over during the entire simulation, and $\varphi_{SR}$ means $\varphi_1$, $\varphi_2$ and $\varphi_3$ must be held simultaneously to indicate the task completes successfully. 
\end{tcolorbox}

\begin{tcolorbox}[title={Running},breakable]

\textbf{Environmental Setup}

The initial position of the quadruped robot is located at (0,0,0) in world coordinates.  A specified forward direction of the robot is along with the positive direction of the x-axis.

The observable states for this task can be found in the natural language input to the LLM (see Appendix ~\ref{anymal_running_prompt}).

The objective metrics for evaluation are defined in the feedback prompt presented in Appendix ~\ref{anymal_running_feedback}, \\

\textbf{Task Specification (STL)}

The success condition for this task is defined as:
\begin{align*}
    & \varphi_1 \equiv \square_{[0.8, 5]} (\text{robot}_{v-x} \geq 2) \\
    & \varphi_2 \equiv \square_{[0, 5]} (|\text{robot}_{pos-y}| \leq 2) \\
    & \varphi_3 \equiv \square_{[0, 5]} (\text{robot}_{pos-z} \geq 0.5) \\
    & \varphi_{SR} \equiv \varphi_2 \wedge \varphi_2 \wedge \varphi_3
\end{align*}

where $\varphi_1$ means that following an initial acceleration phase of 0.8 seconds, the robot must maintain a speed of 2 m/s along the x-axis until the simulation stops at 5 seconds, $\varphi_2$ means that the lateral deviation of the robot should be less than 2 m throughout the simulation, $\varphi_3$ means the robot cannot fall over during the entire simulation, and $\varphi_{SR}$ means $\varphi_1$, $\varphi_2$ and $\varphi_3$ must be held simultaneously to indicate the task completes successfully.

\end{tcolorbox}

\begin{tcolorbox}[title={Walking to Target},breakable]

\textbf{Environmental Setup}

The initial position of the quadruped robot is located at (0,0,0) in world coordinates.  A target position is randomly generated within an area of $(x=[1,3], y=[-1,1],z=0.7)$. 

The observable states for this task can be found in the natural language input to the LLM (see Appendix ~\ref{anymal_velocity_prompt}).

The objective metrics for evaluation are defined in the feedback prompt presented in Appendix ~\ref{anymal_velocity_feedback}, \\

\textbf{Task Specification (STL)}

The success condition for this task is defined as:
\begin{align*}
    & \varphi_1 \equiv \Diamond_{[0, 5]} (\parallel \text{robot}_{pos-xy} - \text{target}_{pos-xy} \parallel \leq 0.5) \\
    & \varphi_2 \equiv \square_{[0, 5]} (\text{robot}_{pos-z} \geq 0.5) \\
    & \varphi_{SR} \equiv \varphi_2 \wedge \varphi_2
\end{align*}

where $\varphi_1$ means that eventually, the distance between the robot and the target point on the x and y axes should be less than 0.5 m, $\varphi_2$ means the robot cannot fall over during the entire simulation, and $\varphi_{SR}$ means $\varphi_1$ and $\varphi_2$  must be held simultaneously to indicate the task completes successfully. 

\end{tcolorbox}

\subsection{Quadcopter}

\begin{tcolorbox}[title={Hovering},breakable]

\textbf{Environmental Setup}

The initial position of the quadcopter is located at (0,0,1) in world coordinates.  A target position is randomly generated within a squared area of $(x=[-2,2], y=[=2,2],z=2)$. 

The observable states for this task can be found in the natural language input to the LLM (see Appendix ~\ref{crazyflie_hovering_prompt}).

The objective metrics for evaluation are defined in the feedback prompt presented in Appendix ~\ref{crazyflie_hovering_feedback}, \\

\textbf{Task Specification (STL)}

The success condition for this task is defined as:
\begin{align*}
    & \varphi_1 \equiv \Diamond_{[0, 30]} (\parallel \text{robot}_{pos} - \text{target}_{pos} \parallel \leq 0.2) \\
    & \varphi_2 \equiv \square_{[0, 30]} ((\text{robot}_{pos-z} \geq 0.5) \wedge (\text{robot}_{pos-z} \leq 5.0)) \\
    & \varphi_{SR} \equiv \varphi_2 \wedge \varphi_2
\end{align*}

where $\varphi_1$ means that eventually, the distance between the quadcopter and the target point should be less than 0.2 m, $\varphi_2$ means the quadcopter cannot fly too high or too low during the entire simulation, and $\varphi_{SR}$ means $\varphi_1$ and $\varphi_2$  must be held simultaneously to indicate the task completes successfully. 
\end{tcolorbox}

\begin{tcolorbox}[title={Wind Field},breakable]

\textbf{Environmental Setup}

The initial position of the quadcopter is located at (0,0,1) in world coordinates.  A target position is randomly generated within a squared area of $(x=[1,2], y=[=2,2],z=2)$. A wind field is defined within an area of $(0.2 < x <0.6)$ with a force of 0.1 N.

The observable states for this task can be found in the natural language input to the LLM (see Appendix ~\ref{crazyflie_wind_prompt}).

The objective metrics for evaluation are defined in the feedback prompt presented in Appendix ~\ref{crazyflie_wind_feedback}, \\

\textbf{Task Specification (STL)}

The success condition for this task is defined as:
\begin{align*}
    & \varphi_1 \equiv \Diamond_{[0, 30]} (\parallel \text{robot}_{pos} - \text{target}_{pos} \parallel \leq 0.2) \\
    & \varphi_2 \equiv \square_{[0, 30]} ((\text{robot}_{pos-z} \geq 0.5) \wedge (\text{robot}_{pos-z} \leq 5.0)) \\
    & \varphi_{SR} \equiv \varphi_2 \wedge \varphi_2
\end{align*}

where $\varphi_1$ means that eventually, the distance between the quadcopter and the target point should be less than 0.2 m, $\varphi_2$ means the quadcopter cannot fly too high or too low during the entire simulation, and $\varphi_{SR}$ means $\varphi_1$ and $\varphi_2$  must be held simultaneously to indicate the task completes successfully. 

\end{tcolorbox}

\begin{tcolorbox}[title={Velocity Tracking},breakable]
\textbf{Environmental Setup}

The initial position of the quadcopter is located at (0,0,1) in world coordinates.  A target velocity is randomly defined in a range of $(v_x = [0.5,1], v_y=[0.5,1], v_z=0)$.

The observable states for this task can be found in the natural language input to the LLM (see Appendix ~\ref{crazyflie_velocity_prompt}).

The objective metrics for evaluation are defined in the feedback prompt presented in Appendix ~\ref{crazyflie_velocity_feedback}, \\

\textbf{Task Specification (STL)}

The success condition for this task is defined as:
\begin{align*}
    & \varphi_1 \equiv \square_{[0, 30]} (\parallel \text{robot}_{v} - \text{target}_{v} \parallel \leq 0.2) \\
    & \varphi_2 \equiv \square_{[0, 30]} (|\text{robot}_{pos-z} - 1.0| \leq 5.0)) \\
    & \varphi_3 \equiv \square_{[0, 30]} ((\text{robot}_{pos-z} \geq 0.5) \wedge (\text{robot}_{pos-z} \leq 5.0)) \\
    & \varphi_{SR} \equiv \varphi_2 \wedge \varphi_2 \wedge \varphi_3
\end{align*}

where $\varphi_1$ means that eventually, the distance between the quadcopter and the target point should be less than 0.2 m, $\varphi_2$ means the quadcopter should maintain the initial height within a deviation of 0.2 m throughout the simulation, $\varphi_3$ means the quadcopter cannot fly too high or too low during the entire simulation, and $\varphi_{SR}$ means $\varphi_1$, $\varphi_2$ and $\varphi_3$  must be held simultaneously to indicate the task completes successfully.

\end{tcolorbox}

\section{Experimental Log}
\label{sec.appendix_experimental_log}

In this section,  we present the reward functions designed by LLM in different self-refinement iterations and the corresponding evaluation results.
We also give the details about the manually designed reward functions.

\subsection{Robotic Manipulator}

\subsubsection{Ball Catching}
\begin{tcolorbox}[title={Iteration 0}, breakable]
\textbf{Reward Function:}

\begin{python}
def reward_function(ball_pos, ball_vel, tray_pos, tray_rot):
    # Constants
    DISTANCE_WEIGHT = 1.0
    VELOCITY_WEIGHT = 1.0
    ROTATION_WEIGHT = 1.0

    # Calculate the distance between the ball and the tray
    distance = np.linalg.norm(ball_pos - tray_pos)

    # Calculate the velocity of the ball
    velocity = np.linalg.norm(ball_vel)

    # Calculate the rotation of the tray
    rotation = np.linalg.norm(tray_rot)

    # Reward for catching the ball
    catch_reward = DISTANCE_WEIGHT * (1.0 / (1.0 + distance))

    # Reward for low velocity
    velocity_reward = VELOCITY_WEIGHT * (1.0 / (1.0 + velocity))

    # Reward for stable tray
    rotation_reward = ROTATION_WEIGHT * (1.0 / (1.0 + rotation))

    # Total reward
    total_reward = catch_reward + velocity_reward + rotation_reward

    return total_reward
\end{python}

\textbf{Evaluation:}

The performance of the RL agent trained with the designed reward function is good.  \\

The training reward converges after 12000 steps. \\
The average timestep for one episode is 299. \\
The average reward is 251 for one episode. \\

Evaluation results of 100 trials are given below: \\
Goal 1 success rate is 100\%  \\
Goal 2 success rate is 100\%  \\

The average normalized action value is 2.012 \\
The normalized distance between the ball and the container is 0.03 \\
The average distance between the robot and the target position on x-axis is 0.51 \\
The average distance between the robot and the target position on y-axis is 0.05 \\
The average distance between the robot and the target position on z-axis is 0.143 \\
The ball's average linear velocity on the x-axis is -0.002 \\
The ball's average linear velocity on the y-axis is -0.0014 \\
The ball's average linear velocity on the z-axis is -0.004 \\

This reward function is qualified as the final reward function.

\end{tcolorbox}

\begin{tcolorbox}[title={Manual Reward Function},breakable]
\textbf{Reward Function:}
\begin{python}
def reward_function(ball_pos, ball_vel, tray_pos, tray_rot):

    # 1st reward ball to tool center distance
    ball_center_dist = torch.norm(tool_pos - ball_pos, p=2, dim=-1)
    ball_center_XY_dist = torch.norm(tool_pos[:,0:3] - ball_pos[:,0:3], p=2, dim=-1)
    center_dist_reward = 1.0/(1.0+ball_center_dist*100)

    # 2nd reward: ball is unmoved
    norm_ball_linvel = torch.norm(ball_linvels, p=2, dim=-1)  
    ball_vel_reward = 1.0/(1.0+norm_ball_linvel*100)

    # 3rd reward: rotation not too much
    rot_diff = torch.norm(tool_rot - self.default_tool_rot, p=2, dim=-1)
    tool_rot_reward = 1.0/(1.0+rot_diff)

    # action penalty
    action_penalty = torch.sum(self.actions[:,0:7] ** 2, dim=-1)
    action_penalty = 1 - 1.0 / (1.0 + action_penalty)

    # liveness_reward
    liveness_reward = torch.where(ball_center_XY_dist<0.03, torch.ones_like(center_dist_reward), torch.zeros_like(center_dist_reward))

    # final cumulative reward
    final_reward = 1.0*center_dist_reward + 1.0*ball_vel_reward + 0.0*tool_rot_reward + 0.5*liveness_reward - 0.01*action_penalty

\end{python}
\end{tcolorbox}
\subsubsection{Ball Balancing}
\begin{tcolorbox}[title={Iteration 0},breakable]
\textbf{Reward Function:}
\begin{python}
def reward_function(ball_pos, ball_vel, tray_pos, tray_rot):
    # Constants
    DISTANCE_WEIGHT = 1.0
    VELOCITY_WEIGHT = 1.0
    ROTATION_WEIGHT = 1.0

    # Calculate the distance between the ball and the tray
    distance = np.linalg.norm(ball_pos - tray_pos)

    # Calculate the velocity of the ball
    velocity = np.linalg.norm(ball_vel)

    # Calculate the rotation of the tray
    rotation = np.linalg.norm(tray_rot)

    # Reward for catching the ball
    catch_reward = DISTANCE_WEIGHT * (1.0 / (1.0 + distance))

    # Reward for low velocity
    velocity_reward = VELOCITY_WEIGHT * (1.0 / (1.0 + velocity))

    # Reward for stable tray
    rotation_reward = ROTATION_WEIGHT * (1.0 / (1.0 + rotation))

    # Total reward
    total_reward = catch_reward + velocity_reward + rotation_reward

    return total_reward
\end{python}

\textbf{Evaluation:}

The performance of the RL agent trained with the designed reward function is good.  \\

The training reward converges after 8000 steps. \\
The average timestep for one episode is 299. \\
The average reward is 183 for one episode. \\

Evaluation results of 100 trials are given below: \\
Goal 1 success rate is 100\%  \\
Goal 2 success rate is 100\%  \\

The average normalized action value is 2.142 \\
The normalized distance between the ball and the target position is 0.082 \\
The average distance between the robot and the target position on x-axis is 0.031 \\
The average distance between the robot and the target position on y-axis is 0.002 \\
The average distance between the robot and the target position on z-axis is 0.002 \\
The ball's average linear velocity on the x-axis is -0.002 \\
The ball's average linear velocity on the y-axis is 0.032 \\
The ball's average linear velocity on the z-axis is 0.002 \\

This reward function is qualified as the final reward function.
\end{tcolorbox}

\begin{tcolorbox}[title={Manual Reward Function},breakable]
\textbf{Reward Function:}
\begin{python}
def reward_function(ball_pos, ball_vel, tray_pos, tray_rot):

    # 1st reward: ball keeps in the center 
    ball_center_dist_3d = torch.norm(tool_pos - ball_pos, p=2, dim=-1)
    center_dist_reward = 1.0/(1.0+ball_center_dist_3d)

    # 2nd reward: ball unmove 
    norm_ball_linvel = torch.norm(ball_linvels, p=2, dim=-1)  
    ball_vel_reward = 1.0/(1.0+norm_ball_linvel)

    # 3rd reward: rotation not too much
    rot_diff = torch.norm(tool_rot - self.default_tool_rot, p=2, dim=-1)
    tool_rot_reward = 1.0/(1.0+rot_diff)

    # stay alive
    liveness = torch.where(ball_pos[:,2]>0.4, torch.ones_like(ball_pos[:,2]), torch.zeros_like(ball_pos[:,2]))

    # the weight of center_dist_reward and ball_vel_reward should be similar
    final_reward = 10.0*center_dist_reward + 5.0*ball_vel_reward + 1.0*tool_rot_reward + 1.0*liveness

\end{python}
\end{tcolorbox}

\subsubsection{Ball Pushing}
\begin{tcolorbox}[title={Iteration 0}, breakable]
\textbf{Reward Function:}

\begin{python}
def reward_function():
    distance_reward = 1.0 / (1.0 + np.linalg.norm(ball_pos - hole_pos))
    direction_to_hole = (hole_pos - ball_pos) / (np.linalg.norm(hole_pos - ball_pos) + 1e-6)
    velocity_reward = np.dot(ball_vel, direction_to_hole)
    velocity_reward = max(0, velocity_reward)
    gripper_ball_distance_reward = 1.0 / (1.0 + np.linalg.norm(gripper_pos - ball_pos))
    total_reward = 0.4 * distance_reward + 0.4 * velocity_reward + 0.2 * gripper_ball_distance_reward

    return total_reward
\end{python}

\textbf{Evaluation:}

The performance of the RL agent trained with the designed reward function is bad. \\

The training reward converges after 43000 steps. \\
The average timestep for one episode is 298. \\
The average reward is 183 for one episode. \\

The average normalized action value is 2.012 \\
The normalized distance between the ball and the target hole is 1.7 \\
The average distance between the ball and the target hole on the x-axis is 0.51 \\
The average distance between the ball and the target hole on the y-axis is 0.05 \\
The average distance between the ball and the target hole on the z-axis is 0.143 \\
The ball's average linear velocity on the x-axis is -0.002 \\
The ball's average linear velocity on the y-axis is -0.0014 \\
The ball's average linear velocity on the z-axis is -0.004 \\

Evaluation results of 100 trials are given below: \\
Goal 1 success rate is 0\% \\

Redesign the reward function based on the given feedback.

\end{tcolorbox}

\begin{tcolorbox}[title={Iteration 1}, breakable]
\textbf{Reward Function:}

\begin{python}
def reward_function():
    distance_reward = 1.0 / (1.0 + np.linalg.norm(ball_pos - hole_pos))
    direction_to_hole = (hole_pos - ball_pos) / np.linalg.norm(hole_pos - ball_pos)
    velocity_reward = np.dot(ball_vel, direction_to_hole)
    velocity_reward = max(0, velocity_reward)  # Ensure the reward is positive
    gripper_ball_distance_reward = 1.0 / (1.0 + np.linalg.norm(gripper_pos - ball_pos))
    total_reward = 0.5 * distance_reward + 0.4 * velocity_reward - 0.1 * gripper_ball_distance_penalty

    return total_reward
\end{python}

\textbf{Evaluation:}

The performance of the RL agent trained with the designed reward function is bad. \\

The training reward converges after 51000 steps. \\
The average timestep for one episode is 299. \\
The average reward is 194 for one episode. \\

The average normalized action value is 2.209 \\
The normalized distance between the ball and the target position is 1.68 \\
The average distance between the robot and the target position on x-axis is 0.52 \\
The average distance between the robot and the target position on y-axis is 0.049 \\
The average distance between the robot and the target position on z-axis is 0.143 \\
The ball average linear velocity on the x-axis is 0.001 \\
The ball average linear velocity on the y-axis is 0.0005 \\
The ball average linear velocity on the z-axis is -0.003 \\

Evaluation results of 100 trials are given below: \\
Goal 1 success rate is 0\% \\

Redesign the reward function based on the given feedback.

\end{tcolorbox}

\begin{tcolorbox}[title={Iteration 2}, breakable]
\textbf{Reward Function:}

\begin{python}
def reward_function():
    distance_reward = 2.0 / (1.0 + np.linalg.norm(ball_pos - hole_pos))
    direction_to_hole = (hole_pos - ball_pos) / np.linalg.norm(hole_pos - ball_pos)
    velocity_reward = np.dot(ball_vel, direction_to_hole)
    velocity_reward = max(0, velocity_reward)  # Ensure the reward is positive
    gripper_ball_distance_reward = 1.0 / (1.0 + np.linalg.norm(gripper_pos - ball_pos))
    total_reward = 0.6 * distance_reward + 0.3 * velocity_reward - 0.1 * gripper_ball_distance_penalty

    return total_reward
\end{python}

\textbf{Evaluation:}

The performance of the RL agent trained with the designed reward function is bad. \\

The training reward converges after 39000 steps. \\
The average timestep for one episode is 297. \\
The average reward is 205 for one episode. \\

The average normalized action value is 1.912 \\
The normalized distance between the ball and the target position is 1.42 \\
The average distance between the robot and the target position on x-axis is 0.41 \\
The average distance between the robot and the target position on y-axis is 0.051 \\
The average distance between the robot and the target position on z-axis is 0.141 \\
The ball average linear velocity on the x-axis is -0.001 \\
The ball average linear velocity on the y-axis is -0.002 \\
The ball average linear velocity on the z-axis is 0.0013 \\

Evaluation results of 100 trials are given below: \\
Goal 1 success rate is 10\% \\

Redesign the reward function based on the given feedback.

\end{tcolorbox}

\begin{tcolorbox}[title={Iteration 3}, breakable]
\textbf{Reward Function:}

\begin{python}
def reward_function():
    distance_reward = 3.0 / (1.0 + np.linalg.norm(ball_pos - hole_pos))
    direction_to_hole = (hole_pos - ball_pos) / np.linalg.norm(hole_pos - ball_pos)
    velocity_reward = np.dot(ball_vel, direction_to_hole)
    velocity_reward = max(0, velocity_reward)  # Ensure the reward is positive
    gripper_ball_distance_reward = 1.0 / (1.0 + np.linalg.norm(gripper_pos - ball_pos))
    total_reward = 0.7 * distance_reward + 0.2 * velocity_reward - 0.1 * gripper_ball_distance_penalty

    return total_reward
\end{python}

\textbf{Evaluation:}

The performance of the RL agent trained with the designed reward function is bad. \\

The training reward converges after 10100 steps. \\
The average timestep for one episode is 291. \\
The average reward is 293 for one episode. \\

The average normalized action value is 2.16 \\
The normalized distance between the ball and the target position is 0.23 \\
The average distance between the robot and the target position on x-axis is 0.194 \\
The average distance between the robot and the target position on y-axis is 0.065 \\
The average distance between the robot and the target position on z-axis is 0.065 \\
The ball average linear velocity on the x-axis is -0.085 \\
The ball average linear velocity on the y-axis is -0.011 \\
The ball average linear velocity on the z-axis is -0.04 \\

Evaluation results of 100 trials are given below: \\
Goal 1 success rate is 82\% \\

Redesign the reward function based on the given feedback.

\end{tcolorbox}

\begin{tcolorbox}[title={Iteration 4}, breakable]
\textbf{Reward Function:}

\begin{python}
def reward_function():
    distance_reward = 4.0 / (1.0 + np.linalg.norm(ball_pos - hole_pos))
    direction_to_hole = (hole_pos - ball_pos) / np.linalg.norm(hole_pos - ball_pos)
    velocity_reward = np.dot(ball_vel, direction_to_hole)
    velocity_reward = max(0, velocity_reward)  # Ensure the reward is positive
    gripper_ball_distance_reward = 1.0 / (1.0 + np.linalg.norm(gripper_pos - ball_pos))
    total_reward = 0.7 * distance_reward + 0.2 * velocity_reward - 0.1 * gripper_ball_distance_penalty

    return total_reward
\end{python}

\textbf{Evaluation:}

The performance of the RL agent trained with the designed reward function is bad. \\

The training reward converges after 9940 steps. \\
The average timestep for one episode is 293. \\
The average reward is 341 for one episode. \\

The average normalized action value is 2.11 \\
The normalized distance between the ball and the target position is 0.19 \\
The average distance between the robot and the target position on x-axis is 0.161 \\
The average distance between the robot and the target position on y-axis is 0.069 \\
The average distance between the robot and the target position on z-axis is 0.068 \\
The ball average linear velocity on the x-axis is -0.109 \\
The ball average linear velocity on the y-axis is -0.016 \\
The ball average linear velocity on the z-axis is -0.0374 \\

Evaluation results of 10 trials are given below: \\
Goal 1 success rate is 90\% \\

Redesign the reward function based on the given feedback.

\end{tcolorbox}

\begin{tcolorbox}[title={Iteration 5}, breakable]
\textbf{Reward Function:}

\begin{python}
def reward_function():
    distance_reward = 5.0 / (1.0 + np.linalg.norm(ball_pos - hole_pos))
    direction_to_hole = (hole_pos - ball_pos) / np.linalg.norm(hole_pos - ball_pos)
    velocity_reward = np.dot(ball_vel, direction_to_hole)
    velocity_reward = max(0, velocity_reward)  # Ensure the reward is positive
    gripper_ball_distance_reward = 1.0 / (1.0 + np.linalg.norm(gripper_pos - ball_pos))
    total_reward = 0.8 * distance_reward + 0.1 * velocity_reward - 0.1 * gripper_ball_distance_penalty

    return total_reward
\end{python}

\textbf{Evaluation:}

The performance of the RL agent trained with the designed reward function is bad. \\

The training reward converges after 9940 steps. \\
The average timestep for one episode is 295. \\
The average reward is 365 for one episode. \\

The average normalized action value is 2.16 \\
The normalized distance between the ball and the target position is 0.142 \\
The average distance between the robot and the target position on x-axis is 0.131 \\
The average distance between the robot and the target position on y-axis is 0.0614 \\
The average distance between the robot and the target position on z-axis is 0.067 \\
The ball average linear velocity on the x-axis is -0.101 \\
The ball average linear velocity on the y-axis is -0.011 \\
The ball average linear velocity on the z-axis is -0.004 \\

Evaluation results of 100 trials are given below: \\
Goal 1 success rate is 91\% \\

Redesign the reward function based on the given feedback.

\end{tcolorbox}

\begin{tcolorbox}[title={Manual Reward Function},breakable]
\textbf{Reward Function:}
\begin{python}
def reward_function(hole_pos, ball_pos, finger_pos, ball_init_pos, actions):

    ball_hole_XY_dist = torch.norm(hole_pos[:,0:2] - ball_pos[:,0:2], p=2, dim=-1)
    
    dist_reward = 1-torch.tanh(3*ball_hole_XY_dist)      # regulize the dist_reward in [0,1]
    
    ball_to_init_dist = torch.norm(ball_pos[:,0:2] - ball_init_pos[:,0:2], p=2, dim=-1)
    self.ball_to_init_dist = ball_to_init_dist
    
    finger_ball_dist = torch.norm(finger_pos - ball_pos, p=2, dim=-1)
    finger_ball_reward = 1.0/(1.0+finger_ball_dist**2)

    # 1st penalty: regularization on the actions (summed for each environment)
    action_penalty = torch.sum(self.actions ** 2, dim=-1)
    action_penalty = 1-torch.tanh(action_penalty/2.5)
    
    # 5th penalty if ball is not moved
    ball_unmove_penalty = torch.zeros_like(dist_reward)
    ball_unmove_penalty = torch.where(ball_to_init_dist<0.3, torch.tanh(15*(0.3-ball_to_init_dist)), ball_unmove_penalty)
    
    falling_bonus = torch.where(torch.logical_and(ball_hole_XY_dist < 0.1 , ball_pos[:,2]<0.38), torch.ones_like(dist_reward), torch.zeros_like(dist_reward))
    
    falling_penalty = torch.zeros_like(dist_reward)
    falling_penalty = torch.where(torch.logical_and(ball_hole_XY_dist > 0.001 , ball_pos[:,2]<0.38), falling_penalty+10, falling_penalty)
    
    dist_reward = torch.where(ball_pos[:,0]<hole_pos[:,0], torch.zeros_like(dist_reward), dist_reward)
    
    dist_penalty = torch.tanh(3*ball_hole_XY_dist) 
    
    final_reward = 10.0*dist_reward - 1.0*ball_unmove_penalty + 100.0*falling_bonus - 0.1*action_penalty \
                  - 1.0*falling_penalty + 1.0*finger_ball_reward - 0.1*dist_penalty
    return final_reward
\end{python}
\end{tcolorbox}

\subsection{Quadruped Robot}

\subsubsection{Running}
\begin{tcolorbox}[title={Iteration 0},breakable]
\textbf{Reward Function:}
\begin{python}
def reward_function(robot_pos, robot_rot, robot_linvel, robot_angvel, actions):

    # Reward for moving forward along the x-axis
    reward_x_velocity = robot_linvel[0]
    
    # Reward for staying upright
    reward_upright = 0.0
    if robot_pos[2] >= 0.5:
        reward_upright = 1.0
    
    # Reward for moving in a straight line (not deviating along y-axis)
    reward_y_deviation = 1.0 - abs(robot_linvel[1])
    
    # Reward for not rotating (staying in the same direction)
    reward_rotation = 1.0 - abs(robot_angvel[2])
    
    # Combine all rewards
    total_reward = reward_x_velocity + reward_upright + reward_y_deviation + reward_rotation
    
    return total_reward
\end{python}

\textbf{Evaluation:}

The performance of the RL agent trained with the designed reward function is bad. \\

The training reward converges after 30000 steps. \\
The average timestep for one episode is 283, \\
The average reward is 2900 for one episode. \\

Testing results of 100 trials are given below: \\
Goal 1 success rate is 10\% \\
Goal 2 success rate is 82\% \\
Goal 3 success rate is 90\% \\

The average normalized action value is 2.36 \\
The robot's average linear velocity on the x-axis is 2.52 \\
The robot's average linear velocity on the y-axis is -0.11 \\
The robot's average linear velocity on the z-axis is -0.23 \\
The robot's average position on the z-axis is 0.61 \\
The robot's average angular velocity on the x-axis is -0.07 \\
The robot's average angular velocity on the y-axis is -0.01 \\
The robot's average angular velocity on the z-axis is -0.01 \\

Redesign the reward function based on the given feedback.
\end{tcolorbox}

\begin{tcolorbox}[title={Iteration 1},breakable]
\textbf{Reward Function:}
\begin{python}
def reward_function(robot_pos, robot_rot, robot_linvel, robot_angvel, actions):
    # Reward for moving forward along the x-axis
    reward_x_velocity = robot_linvel[0] * 2.0  # Increase the weight for moving along x-axis
    
    # Reward for staying upright
    reward_upright = 0.0
    if robot_pos[2] >= 0.5:
        reward_upright = 1.0
    
    # Penalty for moving in y and z directions
    reward_yz_movement = 1.0 - abs(robot_linvel[1]) - abs(robot_linvel[2])
    
    # Reward for not rotating (staying in the same direction)
    reward_rotation = 1.0 - abs(robot_angvel[2])
    
    # Reward for maintaining low action values
    reward_low_action = 1.0 - (sum(abs(actions)) / len(actions)) / 2.36  # Normalize by the average action value
    
    # Combine all rewards
    total_reward = reward_x_velocity + reward_upright + reward_yz_movement + reward_rotation + reward_low_action
    
    return total_reward
\end{python}

\textbf{Evaluation:}

The performance of the RL agent trained with the designed reward function is bad. \\

The training reward converges after 100000 steps. \\
The average timestep for one episode is 292, \\
The average reward is 3103 for one episode. \\

Testing results of 100 trials are given below: \\
Goal 1 success rate is 90\% \\
Goal 2 success rate is 92\% \\
Goal 3 success rate is 95\% \\

The average normalized action value is 2.74 \\
The robot's average linear velocity on the x-axis is 3.23 \\
The robot's average linear velocity on the y-axis is -0.12 \\
The robot's average linear velocity on the z-axis is -0.22 \\
The robot's average position on the z-axis is 0.59 \\
The robot's average angular velocity on the x-axis is -0.1 \\
The robot's average angular velocity on the y-axis is -0.02 \\
The robot's average angular velocity on the z-axis is -0.01 \\

Redesign the reward function based on the given feedback. \\
\end{tcolorbox}

\begin{tcolorbox}[title={Iteration 2},breakable]
\textbf{Reward Function:}
\begin{python}
def reward_function(robot_pos, robot_rot, robot_linvel, robot_angvel, actions):
    # Reward for moving forward along the x-axis
    reward_x_velocity = robot_linvel[0] * 2.0  # Increase the weight for moving along x-axis
    
    # Reward for staying upright
    reward_upright = 0.0
    if robot_pos[2] >= 0.5:
        reward_upright = 1.0
    
    # Penalty for moving in y and z directions
    reward_yz_movement = 1.0 - 2.0 * (abs(robot_linvel[1]) + abs(robot_linvel[2]))  # Increase the penalty for moving in y and z directions
    
    # Reward for not rotating (staying in the same direction)
    reward_rotation = 1.0 - abs(robot_angvel[2])
    
    # Reward for maintaining low action values
    reward_low_action = 1.0 - (sum(abs(actions)) / len(actions)) / 2.74  # Normalize by the new average action value
    
    # Combine all rewards
    total_reward = reward_x_velocity + reward_upright + reward_yz_movement + reward_rotation + reward_low_action
    
    return total_reward
\end{python}

\textbf{Evaluation:}

The performance of the RL agent trained with the designed reward function is good. \\

The training reward converges after 130000 steps. \\
The average timestep for one episode is 299, \\
The average reward is 3274 for one episode. \\

Testing results of 100 trials are given below:

Goal 1 success rate is 98\%  \\
Goal 2 success rate is 99\%  \\
Goal 3 success rate is 100\% \\

The average normalized action value is 2.67 \\
The robot's average linear velocity on the x-axis is 3.76 \\
The robot's average linear velocity on the y-axis is -0.11 \\
The robot's average linear velocity on the z-axis is -0.21 \\
The robot's average position on the z-axis is 0.604 \\
The robot's average angular velocity on the x-axis is -0.04 \\
The robot's average angular velocity on the y-axis is -0.01 \\
The robot's average angular velocity on the z-axis is 0.01 \\

This reward function is qualified as the final reward function.
\end{tcolorbox}

\begin{tcolorbox}[title={Manual Reward Function},breakable]
\textbf{Reward Function:}
\begin{python}
def reward_function(robot_pos, robot_rot, robot_linvel, robot_angvel, actions):
    # velocity bonus for moving forward
    velocity_reward = 1.5*robot_linvel[:,0]

    # balance bonus
    balance_reward = torch.where(robot_pos[:,2] >=0.5, 1.0, 0.0)

    # deviation penalty on y-axis
    dev_penalty = 0.5*robot_pos[:,1]/2.0

    # rotation penalty
    rot_penalty = 0.1*robot_rot[:,1]

    final_reward = velocity_reward + balance_reward - dev_penalty - rot_penalty
    
\end{python}
\end{tcolorbox}

\subsubsection{Velocity Tracking}
\begin{tcolorbox}[title={Iteration 0}, breakable]
\textbf{Reward Function:}

\begin{python}
def reward_function():
    # Constants
    target_speed = 3.0
    min_height = 0.5
    smoothness_weight = 0.1
    stability_weight = 0.1
    speed_weight = 1.0

    # Reward for moving forward at the target speed
    speed_reward = speed_weight * min(target_speed, robot_linvel[0])

    # Reward for not falling over
    height_reward = 0 if robot_pos[2] < min_height else 1

    # Reward for walking smoothly (not jumping or rotating in the air)
    smoothness_reward = smoothness_weight * (1 - abs(robot_linvel[2]) - abs(robot_angvel[0]) - abs(robot_angvel[1]) - abs(robot_angvel[2]))

    # Reward for walking stably (not rotating)
    stability_reward = stability_weight * (1 - abs(robot_rot[0]) - abs(robot_rot[1]) - abs(robot_rot[2]) - abs(robot_rot[3]))

    # Total reward
    total_reward = speed_reward + height_reward + smoothness_reward + stability_reward

    return total_reward
\end{python}

\textbf{Evaluation:}

The performance of the RL agent trained with the designed reward function is bad. \\

The training reward converges after 80000 steps. \\
The average timestep for one episode is 300. \\
The average reward is -521 for one episode. \\

The average normalized action value is 2.79 \\
The robot's average linear velocity on the x-axis is 0.12 \\
The robot's average linear velocity on the y-axis is 0.05 \\
The robot's average linear velocity on the z-axis is -0.027 \\
The robot's average position on the z-axis is 0.227 \\
The robot's average angular velocity on the x-axis is -0.08 \\
The robot's average angular velocity on the y-axis is 0.075 \\
The robot's average angular velocity on the z-axis is 0.343 \\

Evaluation results of 100 trials are given below: \\
Goal 1 success rate is 2\% \\
Goal 2 success rate is 10\% \\
Goal 3 success rate is 40\% \\

Redesign the reward function based on the given feedback.

\end{tcolorbox}

\begin{tcolorbox}[title={Iteration 1}, breakable]
\textbf{Reward Function:}

\begin{python}
def reward_function(robot_pos, robot_rot, robot_linvel, robot_angvel, actions):
    # Constants
    target_speed = 1.0
    min_height = 0.5
    smoothness_weight = 0.1
    stability_weight = 0.1
    speed_weight = 1.0

    # Reward for moving forward at the target speed
    speed_reward = speed_weight * (1 - abs(robot_linvel[0] - target_speed) / target_speed)

    # Reward for not falling over
    height_reward = 0 if robot_pos[2] < min_height else 1

    # Reward for walking smoothly (not jumping or rotating in the air)
    smoothness_reward = smoothness_weight * (1 - abs(robot_linvel[2]) - abs(robot_angvel[0]) - abs(robot_angvel[1]) - abs(robot_angvel[2]))

    # Reward for walking stably (not rotating)
    stability_reward = stability_weight * (1 - abs(robot_rot[0]) - abs(robot_rot[1]) - abs(robot_rot[2]) - abs(robot_rot[3]))

    # Total reward
    total_reward = speed_reward + height_reward + smoothness_reward + stability_reward

    return total_reward
\end{python}

\textbf{Evaluation:}

The performance of the RL agent trained with the designed reward function is bad. \\

The training reward converges after 16000 steps. \\
The average timestep for one episode is 588. \\
The average reward is 1109 for one episode. \\

TThe average normalized action value is 2.36 \\
The robot's average linear velocity on the x-axis is 0.98 \\
The robot's average linear velocity on the y-axis is -0.0515 \\
The robot's average linear velocity on the z-axis is 0.03 \\
The robot's average position on the z-axis is 0.61 \\
The robot's average angular velocity on the x-axis is 0.005 \\
The robot's average angular velocity on the y-axis is -0.01 \\
The robot's average angular velocity on the z-axis is -0.02 \\

Evaluation results of 100 trials are given below: \\
Goal 1 success rate is 50\% \\
Goal 2 success rate is 90\% \\
Goal 3 success rate is 90\% \\

Redesign the reward function based on the given feedback.

\end{tcolorbox}

\begin{tcolorbox}[title={Iteration 2}, breakable]
\textbf{Reward Function:}

\begin{python}
def reward_function(robot_pos, robot_rot, robot_linvel, robot_angvel, actions):
    # Constants
    target_speed = 1.0
    min_height = 0.5
    smoothness_weight = 0.1
    stability_weight = 0.1
    speed_weight = 1.0
    straightness_weight = 0.2

    # Reward for moving forward at the target speed
    speed_reward = speed_weight * (1 - abs(robot_linvel[0] - target_speed) / target_speed)

    # Reward for not falling over
    height_reward = 0 if robot_pos[2] < min_height else 1

    # Reward for walking smoothly (not jumping or rotating in the air)
    smoothness_reward = smoothness_weight * (1 - abs(robot_linvel[2]) - abs(robot_angvel[0]) - abs(robot_angvel[1]) - abs(robot_angvel[2]))

    # Reward for walking stably (not rotating)
    stability_reward = stability_weight * (1 - abs(robot_rot[0]) - abs(robot_rot[1]) - abs(robot_rot[2]) - abs(robot_rot[3]))

    # Reward for moving straight along the x-axis
    straightness_reward = straightness_weight * (1 - abs(robot_linvel[1]))

    # Total reward
    total_reward = speed_reward + height_reward + smoothness_reward + stability_reward + straightness_reward

    return total_reward
\end{python}

\textbf{Evaluation:}

The performance of the RL agent trained with the designed reward function is bad. \\

The training reward converges after 20000 steps. \\
The average timestep for one episode is 595. \\
The average reward is 1220 for one episode. \\

The average normalized action value is 2.432 \\
The robot's average linear velocity on the x-axis is 1.932 \\
The robot's average linear velocity on the y-axis is 0.03 \\
The robot's average linear velocity on the z-axis is -0.12 \\
The robot's average position on the z-axis is 0.59 \\
The robot's average angular velocity on the x-axis is -0.02 \\
The robot's average angular velocity on the y-axis is -0.001 \\
The robot's average angular velocity on the z-axis is -0.008 \\

Evaluation results of 100 trials are given below: \\
Goal 1 success rate is 80\% \\
Goal 2 success rate is 100\% \\
Goal 3 success rate is 100\% \\

Redesign the reward function based on the given feedback.

\end{tcolorbox}

\begin{tcolorbox}[title={Iteration 3}, breakable]
\textbf{Reward Function:}

\begin{python}
def reward_function(robot_pos, robot_rot,     # Constants
    target_speed = 2.0
    min_height = 0.5
    smoothness_weight = 0.1
    stability_weight = 0.1
    speed_weight = 1.5
    straightness_weight = 0.2

    # Reward for moving forward at the target speed
    speed_reward = speed_weight * (1 - abs(robot_linvel[0] - target_speed) / target_speed)

    # Reward for not falling over
    height_reward = 0 if robot_pos[2] < min_height else 1

    # Reward for walking smoothly (not jumping or rotating in the air)
    smoothness_reward = smoothness_weight * (1 - abs(robot_linvel[2]) - abs(robot_angvel[0]) - abs(robot_angvel[1]) - abs(robot_angvel[2]))

    # Reward for walking stably (not rotating)
    stability_reward = stability_weight * (1 - abs(robot_rot[0]) - abs(robot_rot[1]) - abs(robot_rot[2]) - abs(robot_rot[3]))

    # Reward for moving straight along the x-axis
    straightness_reward = straightness_weight * (1 - abs(robot_linvel[1]))

    # Total reward
    total_reward = speed_reward + height_reward + smoothness_reward + stability_reward + straightness_reward

    return total_reward
\end{python}

\textbf{Evaluation:}

The performance of the RL agent trained with the designed reward function is good. \\

The training reward converges after 24000 steps. \\
The average timestep for one episode is 598. \\
The average reward is 1450 for one episode. \\

The average normalized action value is 2.021 \\
The robot's average linear velocity on the x-axis is 1.892 \\
The robot's average linear velocity on the y-axis is 0.021\\
The robot's average linear velocity on the z-axis is -0.119 \\
The robot's average position on the z-axis is 0.59 \\
The robot's average angular velocity on the x-axis is -0.005 \\
The robot's average angular velocity on the y-axis is -0.002 \\
The robot's average angular velocity on the z-axis is -0.006 \\

Evaluation results of 100 trials are given below: \\
Goal 1 success rate is 96\% \\
Goal 2 success rate is 100\% \\
Goal 3 success rate is 100\% \\

Redesign the reward function based on the given feedback.

\end{tcolorbox}

\begin{tcolorbox}[title={Manual Reward Function},breakable]
\textbf{Reward Function:}
\begin{python}
def reward_function (robot_pos , robot_rot , robot_linvel, robot_angvel , actions ) :

    speed_dev = torch.normtorch.norm(robot_linvel[:,0:2] - target_seppd[:,0:2], p=2, dim=-1)
    
    velocity_reward = 2.0* 1.0/(1.0+speed_dev)
    
    balance_reward = 1.0*torch.where(robot_pos[:,2] >= 0.05, 1.0, 0.0)
    
    rot_penalty = 0.1*robot_rot[:,1]
    
    final_reward = velocity_reward + balance_reward - rot_penalty
    
    return final_reward

\end{python}
\end{tcolorbox}

\subsubsection{Walking To Target}
\begin{tcolorbox}[title={Iteration 0}, breakable]
\textbf{Reward Function:}

\begin{python}
def reward_function(robot_pos, robot_rot, robot_linvel, robot_angvel, actions, target_pos):
    # Constants
    alpha = 1.0
    beta = 0.1
    gamma = 0.01
    delta = 0.001

    # Distance to target
    distance_to_target = np.linalg.norm(robot_pos[:2] - target_pos[:2])
    reward_distance = alpha * (1.0 / (1.0 + distance_to_target))

    # Stability reward
    reward_stability = beta * (1.0 if robot_pos[2] >= 0.5 else 0.0)

    # Velocity reward
    reward_velocity = gamma * (1.0 / (1.0 + np.linalg.norm(robot_linvel)))

    # Angular velocity reward
    reward_angular_velocity = delta * (1.0 / (1.0 + np.linalg.norm(robot_angvel)))

    # Total reward
    reward = reward_distance + reward_stability + reward_velocity + reward_angular_velocity

    return reward
\end{python}

\textbf{Evaluation:}

The performance of the RL agent trained with the designed reward function is bad. \\

The training reward converges after 30000 steps. \\
The average timestep for one episode is 568. \\
The average reward is 451 for one episode. \\

The average normalized distance distance between the robot and the target is 8.44 \\
The average normalized action value is 2.29 \\
The robot's average linear velocity on the x-axis is -0.07 \\
The robot's average linear velocity on the y-axis is -0.01 \\
The robot's average linear velocity on the z-axis is 0.03 \\
The robot's average position on the z-axis is 0.57 \\
The robot's average angular velocity on the x-axis is 0.006 \\
The robot's average angular velocity on the y-axis is 0.005 \\
The robot's average angular velocity on the z-axis is -0.004 \\

Evaluation results of 100 trials are given below: \\
Goal 1 success rate is 0\% \\
Goal 2 success rate is 90\% \\

Redesign the reward function based on the given feedback.

\end{tcolorbox}

\begin{tcolorbox}[title={Iteration 1}, breakable]
\textbf{Reward Function:}

\begin{python}
def reward_function(robot_pos, robot_rot, robot_linvel, robot_angvel, actions, target_pos):
    # Constants
    alpha = 2.0
    beta = 0.1
    gamma = 0.01
    delta = 0.001

    # Distance to target
    distance_to_target = np.linalg.norm(robot_pos[:2] - target_pos[:2])
    reward_distance = alpha * (1.0 / (1.0 + distance_to_target))

    # Stability reward
    reward_stability = beta * (1.0 if robot_pos[2] >= 0.5 else 0.0)

    # Velocity reward
    reward_velocity = gamma * (1.0 / (1.0 + np.linalg.norm(robot_linvel)))

    # Angular velocity reward
    reward_angular_velocity = delta * (1.0 / (1.0 + np.linalg.norm(robot_angvel)))

    # Total reward
    reward = reward_distance + reward_stability + reward_velocity + reward_angular_velocity

    return reward
\end{python}

\textbf{Evaluation:}

The performance of the RL agent trained with the designed reward function is bad. \\

The training reward converges after 42000 steps. \\
The average timestep for one episode is 598. \\
The average reward is 531 for one episode. \\

The average normalized distance  between the robot and the target is 4.12 \\
The average normalized action value is 2.15 \\
The robot's average linear velocity on the x-axis is -0.234 \\
The robot's average linear velocity on the y-axis is 0.122 \\
The robot's average linear velocity on the z-axis is 0.003 \\
The robot's average position on the z-axis is 0.521 \\
The robot's average angular velocity on the x-axis is -0.004 \\
The robot's average angular velocity on the y-axis is -0.001 \\
The robot's average angular velocity on the z-axis is 0.031 \\

Evaluation results of 100 trials are given below: \\
Goal 1 success rate is 10\% \\
Goal 2 success rate is 100\% \\

Redesign the reward function based on the given feedback.

\end{tcolorbox}

\begin{tcolorbox}[title={Iteration 2}, breakable]
\textbf{Reward Function:}

\begin{python}
def reward_function(robot_pos, robot_rot, robot_linvel, robot_angvel, actions, target_pos):
    # Constants
    alpha = 3.0  # Further increased weight for distance reward
    beta = 0.1
    gamma = 0.5
    delta = 0.001
    epsilon = 0.01  # Weight for action reward

    # Distance to target
    distance_to_target = np.linalg.norm(robot_pos[:2] - target_pos[:2])
    reward_distance = alpha * (1.0 / (1.0 + distance_to_target))

    # Stability reward
    reward_stability = beta * (1.0 if robot_pos[2] >= 0.5 else 0.0)

    # Velocity reward
    reward_velocity = gamma * (1.0 / (1.0 + np.linalg.norm(robot_linvel)))

    # Angular velocity reward
    reward_angular_velocity = delta * (1.0 / (1.0 + np.linalg.norm(robot_angvel)))

    # Action reward
    reward_action = epsilon * (1.0 / (1.0 + np.linalg.norm(actions)))

    # Total reward
    reward = reward_distance + reward_stability + reward_velocity + reward_angular_velocity + reward_action

    return reward
\end{python}

\textbf{Evaluation:}

The performance of the RL agent trained with the designed reward function is bad. \\

The training reward converges after 38000 steps. \\
The average timestep for one episode is 599. \\
The average reward is 583 for one episode. \\

The average normalized distance  between the robot and the target is 0.8 \\
The average normalized action value is 2.15 \\
The robot's average linear velocity on the x-axis is -0.132 \\
The robot's average linear velocity on the y-axis is 0.202 \\
The robot's average linear velocity on the z-axis is 0.005 \\
The robot's average position on the z-axis is 0.55 \\
The robot's average angular velocity on the x-axis is 0.009 \\
The robot's average angular velocity on the y-axis is 0.001 \\
The robot's average angular velocity on the z-axis is 0.04 \\

Evaluation results of 100 trials are given below: \\
Goal 1 success rate is 73\% \\
Goal 2 success rate is 100\% \\

Redesign the reward function based on the given feedback.

\end{tcolorbox}

\begin{tcolorbox}[title={Iteration 3}, breakable]
\textbf{Reward Function:}

\begin{python}
def reward_function(robot_pos, robot_rot, robot_linvel, robot_angvel, actions, target_pos):
    # Constants
    alpha = 4.0  # Further increased weight for distance reward
    beta = 0.1
    gamma = 0.5
    delta = 0.001
    epsilon = 0.1  # Increased weight for action reward

    # Distance to target
    distance_to_target = np.linalg.norm(robot_pos[:2] - target_pos[:2])
    reward_distance = alpha * (1.0 / (1.0 + distance_to_target))

    # Stability reward
    reward_stability = beta * (1.0 if robot_pos[2] >= 0.5 else 0.0)

    # Velocity reward
    reward_velocity = gamma * (1.0 / (1.0 + np.linalg.norm(robot_linvel)))

    # Angular velocity reward
    reward_angular_velocity = delta * (1.0 / (1.0 + np.linalg.norm(robot_angvel)))

    # Action reward
    reward_action = epsilon * (1.0 / (1.0 + np.linalg.norm(actions)))

    # Total reward
    reward = reward_distance + reward_stability + reward_velocity + reward_angular_velocity + reward_action

    return reward
\end{python}

\textbf{Evaluation:}

The performance of the RL agent trained with the designed reward function is bad. \\

The training reward converges after 38000 steps. \\
The average timestep for one episode is 599. \\
The average reward is 583 for one episode. \\

The average normalized distance distance between the robot and the target is 0.74 \\
The average normalized action value is 2.06 \\
The robot's average linear velocity on the x-axis is 0.193 \\
The robot's average linear velocity on the y-axis is -0.301 \\
The robot's average linear velocity on the z-axis is 0.0045 \\
The robot's average position on the z-axis is 0.55 \\
The robot's average angular velocity on the x-axis is -0.004 \\
The robot's average angular velocity on the y-axis is 0.001 \\
The robot's average angular velocity on the z-axis is 0.04 \\

Evaluation results of 100 trials are given below: \\
Goal 1 success rate is 80\% \\
Goal 2 success rate is 100\% \\

Redesign the reward function based on the given feedback.

\end{tcolorbox}

\begin{tcolorbox}[title={Iteration 4}, breakable]
\textbf{Reward Function:}

\begin{python}
def reward_function(robot_pos, robot_rot, robot_linvel, robot_angvel, actions, target_pos):
    # Constants
    alpha = 5.0  # Further increased weight for distance reward
    beta = 0.1
    gamma = 0.5
    delta = 0.001
    epsilon = 0.1  # Increased weight for action reward

    # Distance to target
    distance_to_target = np.linalg.norm(robot_pos[:2] - target_pos[:2])
    reward_distance = alpha * (1.0 / (1.0 + distance_to_target))

    # Stability reward
    reward_stability = beta * (1.0 if robot_pos[2] >= 0.5 else 0.0)

    # Velocity reward
    reward_velocity = gamma * (1.0 / (1.0 + np.linalg.norm(robot_linvel)))

    # Angular velocity reward
    reward_angular_velocity = delta * (1.0 / (1.0 + np.linalg.norm(robot_angvel)))

    # Action reward
    reward_action = epsilon * (1.0 / (1.0 + np.linalg.norm(actions)))

    # Total reward
    reward = reward_distance + reward_stability + reward_velocity + reward_angular_velocity + reward_action

    return reward
\end{python}

\textbf{Evaluation:}

The performance of the RL agent trained with the designed reward function is bad. \\

The training reward converges after 40000 steps. \\
The average timestep for one episode is 599. \\
The average reward is 592 for one episode. \\

The average normalized distance distance between the robot and the target is 0.72 \\
The average normalized action value is 2.03 \\
The robot's average linear velocity on the x-axis is 0.182 \\
The robot's average linear velocity on the y-axis is -0.291 \\
The robot's average linear velocity on the z-axis is 0.0042 \\
The robot's average position on the z-axis is 0.56 \\
The robot's average angular velocity on the x-axis is 0.0012 \\
The robot's average angular velocity on the y-axis is -0.0006 \\
The robot's average angular velocity on the z-axis is 0.035 \\

Evaluation results of 100 trials are given below: \\
Goal 1 success rate is 80\% \\
Goal 2 success rate is 100\% \\

Redesign the reward function based on the given feedback.

\end{tcolorbox}

\begin{tcolorbox}[title={Iteration 5}, breakable]
\textbf{Reward Function:}

\begin{python}
def reward_function(robot_pos, robot_rot, robot_linvel, robot_angvel, actions, target_pos):
    # Constants
    alpha = 5.0  
    beta = 0.2 # Increased weight for stability reward
    gamma = 0.5
    delta = 0.001
    epsilon = 0.1  

    # Distance to target
    distance_to_target = np.linalg.norm(robot_pos[:2] - target_pos[:2])
    reward_distance = alpha * (1.0 / (1.0 + distance_to_target))

    # Stability reward
    reward_stability = beta * (1.0 if robot_pos[2] >= 0.5 else 0.0)

    # Velocity reward
    reward_velocity = gamma * (1.0 / (1.0 + np.linalg.norm(robot_linvel)))

    # Angular velocity reward
    reward_angular_velocity = delta * (1.0 / (1.0 + np.linalg.norm(robot_angvel)))

    # Action reward
    reward_action = epsilon * (1.0 / (1.0 + np.linalg.norm(actions)))

    # Total reward
    reward = reward_distance + reward_stability + reward_velocity + reward_angular_velocity + reward_action

    return reward
\end{python}

\textbf{Evaluation:}

The performance of the RL agent trained with the designed reward function is bad. \\

The training reward converges after 38000 steps. \\
The average timestep for one episode is 599. \\
The average reward is 583 for one episode. \\

The average normalized distance distance between the robot and the target is 0.69 \\
The average normalized action value is 2.12 \\
The robot's average linear velocity on the x-axis is 0.172 \\
The robot's average linear velocity on the y-axis is -0.311 \\
The robot's average linear velocity on the z-axis is 0.0039 \\
The robot's average position on the z-axis is 0.55 \\
The robot's average angular velocity on the x-axis is -0.0039 \\
The robot's average angular velocity on the y-axis is 0.0009 \\
The robot's average angular velocity on the z-axis is 0.034 \\

Evaluation results of 100 trials are given below: \\
Goal 1 success rate is 85\% \\
Goal 2 success rate is 100\% \\

Redesign the reward function based on the given feedback.

\end{tcolorbox}

\begin{tcolorbox}[title={Manual Reward Function},breakable]
\textbf{Reward Function:}
\begin{python}
reward_function(target_pos, robot_pos, robot_ros, robot_linvel)
	
	robot_target_dist = torch.norm(robot_linvel[:,0:2] - target_pos[:,0:2] , p=2, dim=-1)
	dist_reward = 10 * 1.0/(1.0+robot_target_dist)
	
	height_reward = 2.0 * torch.where(robot_pos[:,2] > 0.5, 1.0, 0.0)

	stability_reward = 0.5 * 1.0/(1.0 + torch.norm(robot_linvel, p=2, dim=-1))
	
	action_penalty = 0.1 * (1.0 - torch.abs(self.actions).sum(-1)/2.0)

	return dist_reward + height_reward + stability_reward - action_penalty
\end{python}
\end{tcolorbox}


\subsection{Quadcopter}

\subsubsection{Hovering}
\begin{tcolorbox}[title={Iteration 0}, breakable]
\textbf{Reward Function:}

\begin{python}
def reward_function(copter_pos, copter_rot, target_pos, copter_angvels, actions):
    # Constants
    Z_UPPER_LIMIT = 3.0
    Z_LOWER_LIMIT = 0.8
    HOVER_BONUS = 1.0
    POSITION_REWARD_SCALE = 1.0
    ANGVEL_REWARD_SCALE = 0.1
    ACTION_REWARD_SCALE = 0.1

    # Calculate distance to target
    distance_to_target = np.linalg.norm(target_pos - copter_pos)

    # Calculate reward for being close to target position
    position_reward = POSITION_REWARD_SCALE / (1 + distance_to_target)

    # Calculate reward for low angular velocity
    angvel_reward = ANGVEL_REWARD_SCALE / (1 + np.linalg.norm(copter_angvels))

    # Calculate reward for small actions
    action_reward = ACTION_REWARD_SCALE / (1 + np.linalg.norm(actions))

    # Check if copter is hovering at target position
    hover_reward = 0.0
    if distance_to_target < 0.1 and Z_LOWER_LIMIT <= copter_pos[2] <= Z_UPPER_LIMIT:
        hover_reward = HOVER_BONUS

    # Calculate total reward
    total_reward = position_reward + angvel_reward + action_reward + hover_reward

    return total_reward
\end{python}

\textbf{Evaluation:}

The performance of the RL agent trained with the designed reward function is bad. \\

The training reward converges after 4300 steps. \\
The average timestep for one episode is 298. \\
The average reward is 183 for one episode. \\

The average normalized action value is 1.038 \\
The normalized distance between the Quadcopter and the target position is 1.7 \\
The robot's average linear velocity on the x-axis is -0.28 \\
The robot's average linear velocity on the y-axis is 0.29 \\
The robot's average linear velocity on the z-axis is 0.20 \\
The robot's average angular velocity on the x-axis is -0.01 \\
The robot's average angular velocity on the y-axis is 0.008 \\
The robot's average angular velocity on the z-axis is -0.02 \\

Evaluation results of 100 trials are given below: \\
Goal 1 success rate is 10\% \\
Goal 2 success rate is 100\% \\

Redesign the reward function based on the given feedback.

\end{tcolorbox}

\begin{tcolorbox}[title={Iteration 1}, breakable]
\textbf{Reward Function:}

\begin{python}
def reward_function(copter_pos, copter_rot, target_pos, copter_angvels, actions):
 def reward_function(copter_pos, copter_rot, target_pos, copter_angvels, actions):
    # Constants
    Z_UPPER_LIMIT = 3.0
    Z_LOWER_LIMIT = 0.8
    HOVER_BONUS = 1.0
    POSITION_REWARD_SCALE = 2.0  # Increase the weight for position reward
    ANGVEL_REWARD_SCALE = 0.1
    ACTION_REWARD_SCALE = 0.5  # Increase the weight for action reward

    # Calculate distance to target
    distance_to_target = np.linalg.norm(target_pos - copter_pos)

    # Calculate reward for being close to target position
    position_reward = POSITION_REWARD_SCALE / (1 + distance_to_target)

    # Calculate reward for low angular velocity
    angvel_reward = ANGVEL_REWARD_SCALE / (1 + np.linalg.norm(copter_angvels))

    # Calculate reward for small actions
    action_reward = ACTION_REWARD_SCALE / (1 + np.linalg.norm(actions))

    # Check if copter is hovering at target position
    hover_reward = 0.0
    if distance_to_target < 0.1 and Z_LOWER_LIMIT <= copter_pos[2] <= Z_UPPER_LIMIT:
        hover_reward = HOVER_BONUS

    # Calculate total reward
    total_reward = position_reward + angvel_reward + action_reward + hover_reward

    return total_reward
\end{python}

\textbf{Evaluation:}

The performance of the RL agent trained with the designed reward function is bad. \\

The training reward converges after 20000 steps. \\
The average timestep for one episode is 298. \\
The average reward is 342 for one episode. \\

The average normalized action value is 0.95 \\
The normalized distance between the quadruped robot and the target position is 0.37 \\
The robot's average linear velocity on the x-axis is 0.01 \\
The robot's average linear velocity on the y-axis is -0.05 \\
The robot's average linear velocity on the z-axis is 0.16 \\
The robot's average angular velocity on the x-axis is -0.003 \\
The robot's average angular velocity on the y-axis is -0.01 \\
The robot's average angular velocity on the z-axis is 0.02 \\

Evaluation results of 100 trials are given below: \\
Goal 1 success rate is 90\% \\
Goal 2 success rate is 100\% \\

Redesign the reward function based on the given feedback.

\end{tcolorbox}

\begin{tcolorbox}[title={Iteration 2}, breakable]
\textbf{Reward Function:}

\begin{python}
def reward_function(copter_pos, copter_rot, target_pos, copter_angvels, actions):
    # Constants
    Z_UPPER_LIMIT = 3.0
    Z_LOWER_LIMIT = 0.8
    HOVER_BONUS = 1.0
    POSITION_REWARD_SCALE = 2.5  # Further increase the weight for position reward
    ANGVEL_REWARD_SCALE = 0.5  # Increase the weight for angular velocity reward
    ACTION_REWARD_SCALE = 0.5

    # Calculate distance to target
    distance_to_target = np.linalg.norm(target_pos - copter_pos)

    # Calculate reward for being close to target position
    position_reward = POSITION_REWARD_SCALE / (1 + distance_to_target)

    # Calculate reward for low angular velocity
    angvel_reward = ANGVEL_REWARD_SCALE / (1 + np.linalg.norm(copter_angvels))

    # Calculate reward for small actions
    action_reward = ACTION_REWARD_SCALE / (1 + np.linalg.norm(actions))

    # Check if copter is hovering at target position
    hover_reward = 0.0
    if distance_to_target < 0.1 and Z_LOWER_LIMIT <= copter_pos[2] <= Z_UPPER_LIMIT:
        hover_reward = HOVER_BONUS

    # Calculate total reward
    total_reward = position_reward + angvel_reward + action_reward + hover_reward

    return total_reward
\end{python}

\textbf{Evaluation:}

The performance of the RL agent trained with the designed reward function is good. \\

The training reward converges after 5600 steps. \\
The average timestep for one episode is 299. \\
The average reward is 243 for one episode. \\

The average normalized action value is 1.005 \\
The normalized distance between the Quadcopter and the target position is 0.32 \\
The robot's average linear velocity on the x-axis is -0.24 \\
The robot's average linear velocity on the y-axis is 0.22 \\
The robot's average linear velocity on the z-axis is 0.13 \\
The robot's average angular velocity on the x-axis is -0.008 \\
The robot's average angular velocity on the y-axis is -0.006 \\
The robot's average angular velocity on the z-axis is 0.018 \\

Evaluation results of 100 trials are given below: \\
Goal 1 success rate is 98\% \\
Goal 2 success rate is 100\% \\

Redesign the reward function based on the given feedback.

\end{tcolorbox}

\begin{tcolorbox}[title={Manual Reward Function},breakable]
\textbf{Reward Function:}
\begin{python}
reward_function(target_pos, robot_pos, robot_ros, robot_linvel, robot_angvel)
	
	robot_target_dist = torch.norm(target_pos - target_pos , p=2, dim=-1)
	dist_reward = 5.0 * 1.0/(1.0+robot_target_dist)
	
	height_reward = 1.0 * torch.where(root_positions[:,2]<=3.0, torch.where(root_positions[:,2]>=0.8, 1.0, 0.0), 0.0)

	angular_penalty = 0.2 * torch.abs(robot_angvel).sum(-1)
	
	action_penalty = 0.1 * (1.0 - torch.abs(self.actions).sum(-1)/2.0)

	return dist_reward + height_reward - angular_penalty - action_penalty

\end{python}
\end{tcolorbox}

\subsubsection{Wind Field}
\begin{tcolorbox}[title={Iteration 0}, breakable]
\textbf{Reward Function:}

\begin{python}
def reward_function(copter_pos, copter_rot, target_pos, copter_angvels, actions):
    # Distance to target
    distance_to_target = np.linalg.norm(copter_pos - target_pos)
    
    # Reward for being close to the target
    reward_close_to_target = 1.0 / (1.0 + distance_to_target)
    
    # Penalty for exceeding height limit or falling below minimum height
    height_penalty = 0
    if copter_pos[2] > 3.0 or copter_pos[2] < 0.8:
        height_penalty = 1.0

    # Reward for maintaining a stable hover
    reward_hover = 1.0 / (1.0 + np.abs(copter_angvels).sum())
    
    # Total reward
    total_reward = reward_close_to_target + reward_hover - height_penalty
    
    return total_reward
\end{python}

\textbf{Evaluation:}

The performance of the RL agent trained with the designed reward function is bad. \\

The training reward converges after 8000 steps. \\
The average timestep for one episode is 497. \\
The average reward is 624 for one episode. \\

The average normalized action value is 1.21 \\
The average distance between the quadruped robot and the target position on x-axis is 1.6 \\
The average distance between the quadruped robot and the target position on y-axis is 0.44 \\
The average distance between the quadruped robot and the target position on z-axis is 1.23 \\
The robot's average linear velocity on the x-axis is -0.11 \\
The robot's average linear velocity on the y-axis is -0.48 \\
The robot's average linear velocity on the z-axis is -0.06 \\
The robot's average angular velocity on the x-axis is 0.01 \\
The robot's average angular velocity on the y-axis is -0.003 \\
The robot's average angular velocity on the z-axis is 0.07 \\

Evaluation results of 100 trials are given below: \\
Goal 1 success rate is 0\% \\
Goal 2 success rate is 100\% \\

Redesign the reward function based on the given feedback.

\end{tcolorbox}

\begin{tcolorbox}[title={Iteration 1}, breakable]
\textbf{Reward Function:}

\begin{python}
def reward_function(copter_pos, copter_rot, target_pos, copter_angvels, actions):
    # Distance to target
    distance_to_target = np.linalg.norm(copter_pos - target_pos)
    
    # Additional penalty for distance on x-axis
    x_distance_penalty = np.abs(copter_pos[0] - target_pos[0])
    
    # Reward for being close to the target
    reward_close_to_target = 1.0 / (1.0 + distance_to_target + x_distance_penalty)
    
    # Penalty for exceeding height limit or falling below minimum height
    height_penalty = 0
    if copter_pos[2] > 3.0 or copter_pos[2] < 0.8:
        height_penalty = 1.0

    # Reward for maintaining a stable hover
    reward_hover = 1.0 / (1.0 + np.abs(copter_angvels).sum())
    
    # Reward for maintaining low velocities
    reward_low_velocity = 1.0 / (1.0 + np.abs(actions).sum())
    
    # Total reward
    total_reward = reward_close_to_target + reward_hover + reward_low_velocity - height_penalty
    
    return total_reward
\end{python}

\textbf{Evaluation:}

The performance of the RL agent trained with the designed reward function is bad. \\

The training reward converges after 6000 steps. \\
The average timestep for one episode is 499. \\
The average reward is 794 for one episode. \\

The average normalized action value is 0.71 \\
The average distance between the quadruped robot and the target position on x-axis is 1.6 \\
The average distance between the quadruped robot and the target position on y-axis is 1.9 \\
The average distance between the quadruped robot and the target position on z-axis is 1.3 \\
The robot's average linear velocity on the x-axis is -0.05 \\
The robot's average linear velocity on the y-axis is -0.37 \\
The robot's average linear velocity on the z-axis is -0.06 \\
The robot's average angular velocity on the x-axis is 0.01 \\
The robot's average angular velocity on the y-axis is -0.003 \\
The robot's average angular velocity on the z-axis is -0.01 \\

Evaluation results of 100 trials are given below: \\
Goal 1 success rate is 0\% \\
Goal 2 success rate is 100\% \\

Redesign the reward function based on the given feedback.

\end{tcolorbox}

\begin{tcolorbox}[title={Iteration 2}, breakable]
\textbf{Reward Function:}

\begin{python}
def reward_function(copter_pos, copter_rot, target_pos, copter_angvels, actions):
    # Distance to target
    distance_to_target = np.linalg.norm(copter_pos - target_pos)
    
    # Additional penalty for distance on x and y axes
    xy_distance_penalty = np.abs(copter_pos[0] - target_pos[0]) + np.abs(copter_pos[1] - target_pos[1])
    
    # Reward for being close to the target
    reward_close_to_target = 1.0 / (1.0 + distance_to_target + xy_distance_penalty)
    
    # Penalty for exceeding height limit or falling below minimum height
    height_penalty = 0
    if copter_pos[2] > 3.0 or copter_pos[2] < 0.8:
        height_penalty = 1.0

    # Reward for maintaining a stable hover
    reward_hover = 1.0 / (1.0 + np.abs(copter_angvels).sum())
    
    # Reward for maintaining low velocities on x and y axes
    reward_low_velocity = 1.0 / (1.0 + np.abs(actions[0]) + np.abs(actions[1]))
    
    # Total reward
    total_reward = reward_close_to_target + reward_hover + reward_low_velocity - height_penalty
    
    return total_reward
\end{python}

\textbf{Evaluation:}

The performance of the RL agent trained with the designed reward function is bad. \\

The training reward converges after 7000 steps. \\
The average timestep for one episode is 494. \\
The average reward is 945 for one episode. \\

The average normalized action value is 0.71 \\
The average distance between the quadruped robot and the target position on x-axis is 1.46 \\
The average distance between the quadruped robot and the target position on y-axis is 0.73 \\
The average distance between the quadruped robot and the target position on z-axis is 1.0 \\
The robot's average linear velocity on the x-axis is -0.05 \\
The robot's average linear velocity on the y-axis is -0.03 \\
The robot's average linear velocity on the z-axis is -0.001 \\
The robot's average angular velocity on the x-axis is 0.003 \\
The robot's average angular velocity on the y-axis is 0.001 \\
The robot's average angular velocity on the z-axis is -0.05 \\

Evaluation results of 100 trials are given below: \\
Goal 1 success rate is 0\% \\
Goal 2 success rate is 100\% \\

Redesign the reward function based on the given feedback.

\end{tcolorbox}

\begin{tcolorbox}[title={Iteration 3}, breakable]
\textbf{Reward Function:}

\begin{python}
def reward_function(copter_pos, copter_rot, target_pos, copter_angvels, actions):
     # Distance to target
    distance_to_target = np.linalg.norm(copter_pos - target_pos)
    
    # Additional penalty for distance on x, y and z axes
    xyz_distance_penalty = np.abs(copter_pos[0] - target_pos[0]) + np.abs(copter_pos[1] - target_pos[1]) + np.abs(copter_pos[2] - target_pos[2])
    
    # Reward for being close to the target
    reward_close_to_target = 1.0 / (1.0 + distance_to_target + xyz_distance_penalty)
    
    # Penalty for exceeding height limit or falling below minimum height
    height_penalty = 0
    if copter_pos[2] > 3.0 or copter_pos[2] < 0.8:
        height_penalty = 1.0

    # Reward for maintaining a stable hover
    reward_hover = 1.0 / (1.0 + np.abs(copter_angvels).sum())
    
    # Reward for maintaining low velocities on x, y and z axes
    reward_low_velocity = 1.0 / (1.0 + np.abs(actions[0]) + np.abs(actions[1]) + np.abs(actions[2]))
    
    # Total reward
    total_reward = reward_close_to_target + reward_hover + reward_low_velocity - height_penalty
    
    return total_reward
\end{python}

\textbf{Evaluation:}

The performance of the RL agent trained with the designed reward function is bad. \\

The training reward converges after 30000 steps. \\
The average timestep for one episode is 480. \\
The average reward is 967 for one episode. \\

The average normalized action value is 1.75 \\
The average distance between the quadruped robot and the target position on x-axis is 0.4 \\
The average distance between the quadruped robot and the target position on y-axis is 0.42 \\
The average distance between the quadruped robot and the target position on z-axis is 0.29 \\
The robot's average linear velocity on the x-axis is 0.2 \\
The robot's average linear velocity on the y-axis is 0.12 \\
The robot's average linear velocity on the z-axis is 0.124 \\
The robot's average angular velocity on the x-axis is -0.06 \\
The robot's average angular velocity on the y-axis is 0.103 \\
The robot's average angular velocity on the z-axis is 0.36 \\

Evaluation results of 100 trials are given below: \\
Goal 1 success rate is 20\% \\
Goal 2 success rate is 90\% \\

Redesign the reward function based on the given feedback.

\end{tcolorbox}

\begin{tcolorbox}[title={Iteration 4}, breakable]
\textbf{Reward Function:}

\begin{python}
def reward_function(copter_pos, copter_rot, target_pos, copter_angvels, actions):
    # Distance to target
    distance_to_target = np.linalg.norm(copter_pos - target_pos)
    
    # Additional penalty for distance on x, y and z axes
    xyz_distance_penalty = np.abs(copter_pos[0] - target_pos[0]) + np.abs(copter_pos[1] - target_pos[1]) + np.abs(copter_pos[2] - target_pos[2])
    
    # Reward for being close to the target
    reward_close_to_target = 1.0 / (1.0 + distance_to_target + xyz_distance_penalty)
    
    # Penalty for exceeding height limit or falling below minimum height
    height_penalty = 0
    if copter_pos[2] > 3.0 or copter_pos[2] < 0.8:
        height_penalty = 1.0

    # Reward for maintaining a stable hover
    reward_hover = 1.0 / (1.0 + np.abs(copter_angvels).sum())
    
    # Reward for taking smaller actions
    reward_small_actions = 1.0 / (1.0 + np.abs(actions).sum())
    
    # Total reward
    total_reward = reward_close_to_target + reward_hover + reward_small_actions - height_penalty
    
    return total_reward
\end{python}

\textbf{Evaluation:}

The performance of the RL agent trained with the designed reward function is bad. \\

The training reward converges after 34000 steps. \\
The average timestep for one episode is 499. \\
The average reward is 1011 for one episode. \\

The average normalized action value is 1.38 \\
The average distance between the quadruped robot and the target position on x-axis is 0.37 \\
The average distance between the quadruped robot and the target position on y-axis is 0.40 \\
The average distance between the quadruped robot and the target position on z-axis is 0.28 \\
The robot's average linear velocity on the x-axis is 0.193 \\
The robot's average linear velocity on the y-axis is 0.123 \\
The robot's average linear velocity on the z-axis is 0.113 \\
The robot's average angular velocity on the x-axis is -0.066 \\
The robot's average angular velocity on the y-axis is 0.121 \\
The robot's average angular velocity on the z-axis is 0.39 \\

Evaluation results of 100 trials are given below: \\
Goal 1 success rate is 100\% \\
Goal 2 success rate is 100\% \\

Redesign the reward function based on the given feedback.

\end{tcolorbox}

\begin{tcolorbox}[title={Manual Reward Function},breakable]
\textbf{Reward Function:}
\begin{python}
def reward_function(target_pos, robot_pos, robot_ros, robot_linvel, robot_angvel)
	
	robot_target_dist = torch.norm(target_pos - target_pos , p=2, dim=-1)
	dist_reward = 5.0 * 1.0/(1.0+robot_target_dist)
	
	height_reward = 1.0 * torch.where(root_positions[:,2]<=3.0, torch.where(root_positions[:,2]>=0.8, 1.0, 0.0), 0.0)

	angular_penalty = 0.2 * torch.abs(robot_angvel).sum(-1)

	wind_reward = torch.where(robot_linvel[:,0]<=0.5, 1.0, 0.0)
	
	action_penalty = 0.1 * (1.0 - torch.abs(self.actions).sum(-1)/2.0)

	return dist_reward + height_reward + wind_reward - angular_penalty - action_penalty

\end{python}
\end{tcolorbox}

\subsubsection{Velocity Tracking}
\begin{tcolorbox}[title={Iteration 0}, breakable]
\textbf{Reward Function:}
\begin{python}
def reward_function(copter_pos, copter_rot, target_vel, copter_angvels, actions, copter_linvels):
    # Constants
    Z_TARGET = 1.0
    VELOCITY_WEIGHT = 1.0
    HEIGHT_WEIGHT = 0.5

    # Calculate the difference between the target velocity and the copter's current velocity
    vel_diff = np.linalg.norm(target_vel - copter_linvels)

    # Calculate the difference between the target height and the copter's current height
    height_diff = np.abs(Z_TARGET - copter_pos[2])

    # Calculate the reward for the velocity and height
    vel_reward = np.exp(-VELOCITY_WEIGHT * vel_diff)
    height_reward = np.exp(-HEIGHT_WEIGHT * height_diff)

    # Combine the rewards
    reward = vel_reward + height_reward

    return reward
\end{python}

\textbf{Evaluation:}

The performance of the RL agent trained with the designed reward function is bad. \\

The training reward converges after 20000 steps. \\
The average timestep for one episode is 295. \\
The average reward is 385 for one episode. \\

The average normalized action value is 1.588 \\
The linear velocity deviation on the x-axis is -2.582 \\
The linear velocity deviation on the y-axis is -2.292 \\
The linear velocity deviation on the z-axis is -0.005 \\
The normalized linear velocity deviation is 3.52 \\
The robot's average position on the z-axis is 0.9 \\
The robot's average angular velocity on the x-axis is -0.003 \\
The robot's average angular velocity on the y-axis is -0.002 \\
The robot's average angular velocity on the z-axis is -0.001 \\

Evaluation results of 100 trials are given below: \\
Goal 1 success rate is 0\% \\
Goal 2 success rate is 50\% \\

Redesign the reward function based on the given feedback.

\end{tcolorbox}

\begin{tcolorbox}[title={Iteration 1}, breakable]
\textbf{Reward Function:}
\begin{python}
def reward_function(copter_pos, copter_rot, target_vel, copter_angvels, actions, copter_linvels):
    # Constants
    Z_TARGET = 1.0
    VELOCITY_WEIGHT = 1.0
    HEIGHT_WEIGHT = 1.0
    ANGVEL_WEIGHT = 0.1

    # Calculate the difference between the target velocity and the copter's current velocity
    vel_diff = np.linalg.norm(target_vel - copter_linvels)

    # Calculate the difference between the target height and the copter's current height
    height_diff = np.abs(Z_TARGET - copter_pos[2])

    # Calculate the magnitude of the angular velocity
    angvel_magnitude = np.linalg.norm(copter_angvels)

    # Calculate the reward for the velocity, height and angular velocity
    vel_reward = np.exp(-VELOCITY_WEIGHT * vel_diff)
    height_reward = np.exp(-HEIGHT_WEIGHT * height_diff)
    angvel_reward = np.exp(-ANGVEL_WEIGHT * angvel_magnitude)

    # Combine the rewards
    reward = vel_reward + height_reward + angvel_reward

    return reward
\end{python}

\textbf{Evaluation:}

The performance of the RL agent trained with the designed reward function is bad. \\

The training reward converges after 15000 steps. \\
The average timestep for one episode is 292. \\
The average reward is 528 for one episode. \\

The average normalized action value is 1.588
The linear velocity deviation on the x-axis is -2.466 \\
The linear velocity deviation on the y-axis is -2.482 \\
The linear velocity deviation on the z-axis is -0.006 \\
The normalized linear velocity deviation is 3.52 \\
The robot's average position on the z-axis is 0.88 \\
The robot's average angular velocity on the x-axis is -0.006 \\
The robot's average angular velocity on the y-axis is -0.009 \\
The robot's average angular velocity on the z-axis is -0.0002 \\

Evaluation results of 100 trials are given below: \\
Goal 1 success rate is 0\% \\
Goal 2 success rate is 90\% \\

Redesign the reward function based on the given feedback.

\end{tcolorbox}

\begin{tcolorbox}[title={Iteration 2}, breakable]
\textbf{Reward Function:}
\begin{python}
def reward_function(copter_pos, copter_rot, target_vel, copter_angvels, actions, copter_linvels):
    # Constants
    Z_TARGET = 1.0
    VELOCITY_WEIGHT = 1.0
    HEIGHT_WEIGHT = 0.5
    ANGVEL_WEIGHT = 0.1

    # Calculate the difference between the target velocity and the copter's current velocity
    vel_diff = np.linalg.norm(target_vel - copter_linvels)

    # Calculate the difference between the target height and the copter's current height
    height_diff = np.abs(Z_TARGET - copter_pos[2])

    # Calculate the magnitude of the angular velocity
    angvel_magnitude = np.linalg.norm(copter_angvels)

    # Calculate the reward for the velocity, height and angular velocity
    vel_reward = np.exp(-VELOCITY_WEIGHT * vel_diff)
    height_reward = np.exp(-HEIGHT_WEIGHT * height_diff)
    angvel_reward = np.exp(-ANGVEL_WEIGHT * angvel_magnitude)

    # Combine the rewards
    reward = vel_reward + height_reward + angvel_reward

    return reward
\end{python}

\textbf{Evaluation:}

The performance of the RL agent trained with the designed reward function is bad. \\

The training reward converges after 40000 steps. \\
The average timestep for one episode is 299. \\
The average reward is 751 for one episode. \\

The average normalized action value is 1.209 \\
The linear velocity deviation on the x-axis is 0.1617 \\
The linear velocity deviation on the y-axis is -0.01 \\
The linear velocity deviation on the z-axis is -0.1636 \\
The normalized linear velocity deviation is 0.48 \\
The robot's average position on the z-axis is 1.171 \\
The robot's average angular velocity on the x-axis is -0.003 \\
The robot's average angular velocity on the y-axis is 0.009 \\
The robot's average angular velocity on the z-axis is -0.006 \\

Evaluation results of 100 trials are given below: \\
Goal 1 success rate is 70\% \\
Goal 2 success rate is 100\% \\

Redesign the reward function based on the given feedback.

\end{tcolorbox}

\begin{tcolorbox}[title={Iteration 3}, breakable]
\textbf{Reward Function:}
\begin{python}
def reward_function(copter_pos, copter_rot, target_vel, copter_angvels, actions, copter_linvels):
    # Constants
    Z_TARGET = 1.0
    VELOCITY_WEIGHT = 1.5
    HEIGHT_WEIGHT = 1.0
    ANGVEL_WEIGHT = 0.1

    # Calculate the difference between the target velocity and the copter's current velocity
    vel_diff = np.linalg.norm(target_vel - copter_linvels)

    # Calculate the difference between the target height and the copter's current height
    height_diff = np.abs(Z_TARGET - copter_pos[2])

    # Calculate the magnitude of the angular velocity
    angvel_magnitude = np.linalg.norm(copter_angvels)

    # Calculate the reward for the velocity, height and angular velocity
    vel_reward = np.exp(-VELOCITY_WEIGHT * vel_diff)
    height_reward = np.exp(-HEIGHT_WEIGHT * height_diff)
    angvel_reward = np.exp(-ANGVEL_WEIGHT * angvel_magnitude)

    # Combine the rewards
    reward = vel_reward + height_reward + angvel_reward
    return reward
\end{python}

\textbf{Evaluation:}

The performance of the RL agent trained with the designed reward function is good. \\

The training reward converges after 43000 steps. \\
The average timestep for one episode is 299. \\
The average reward is 894 for one episode. \\

The average normalized action value is 1.221 \\
The linear velocity deviation on the x-axis is 0.1212 \\
The linear velocity deviation on the y-axis is -0.0093 \\
The linear velocity deviation on the z-axis is -0.1021 \\
The normalized linear velocity deviation is 0.183 \\
The robot's average position on the z-axis is 1.132 \\
The robot's average angular velocity on the x-axis is -0.011 \\
The robot's average angular velocity on the y-axis is 0.008 \\
The robot's average angular velocity on the z-axis is -0.007 \\

Evaluation results of 100 trials are given below: \\
Goal 1 success rate is 99\% \\
Goal 2 success rate is 100\% \\

Redesign the reward function based on the given feedback.

\end{tcolorbox}

\begin{tcolorbox}[title={Manual Reward Function},breakable]
\textbf{Reward Function:}
\begin{python}
def reward_function(target_vel, robot_pos, robot_ros, robot_linvel, robot_angvel)
	
	velocity_dev = torch.norm(target_vel - robot_linvel , p=2, dim=-1)
	velocity_reward = 3.0 * 1.0/(1.0+velocity_dev)
	
	height_dev = torch.abs(robot_pos[:,2] - 1.0)
	height_reward = 0.5 * 1.0/(1.0+height_dev)

	angular_penalty = 0.1 * torch.abs(robot_angvel).sum(-1)
	
	action_penalty = 0.05 * (1.0 - torch.abs(self.actions).sum(-1)/2.0)

	return velocity_reward + height_reward - angular_penalty - action_penalty
\end{python}
\end{tcolorbox}







\end{document}